\documentclass[5p,times]{elsarticle}%
\usepackage{siunitx}
\usepackage[utf8]{inputenc}
\usepackage{comment}
\usepackage{scalerel}
\usepackage{graphicx}
\usepackage{amsmath}
\usepackage{amsthm}
\usepackage{bookmark}
\usepackage{algorithm}
\usepackage{amsmath,amssymb,amsfonts}
\usepackage{algorithmic} 

\usepackage{makecell, multirow}
\usepackage{graphics} % for pdf, bitmapped graphics files
\usepackage{amsmath} % assumes amsmath package installed
\usepackage{amssymb}  % assumes amsmath package installed
\usepackage{caption}
\usepackage{subcaption}
\usepackage{placeins}
\usepackage[english]{babel}
\usepackage{fancyhdr}
\usepackage{lastpage}
\usepackage{blindtext}
\usepackage{xcolor}
\newcommand\algorithmicprocedure{\textbf{procedure}}
\newcommand{\algorithmicendprocedure}{\algorithmicend\ \algorithmicprocedure}
\makeatletter
\newcommand\PROCEDURE[3][default]{%
  \ALC@it
  \algorithmicprocedure\ \textsc{#2}(#3)%
  \ALC@com{#1}%
  \begin{ALC@prc}%
}
\newcommand\ENDPROCEDURE{%
  \end{ALC@prc}%
  \ifthenelse{\boolean{ALC@noend}}{}{%
    \ALC@it\algorithmicendprocedure
  }%
}
\newenvironment{ALC@prc}{\begin{ALC@g}}{\end{ALC@g}}
\makeatother

\def\RE#1{{\color{orange} {{#1}}}}

\begin{document}
\begin{frontmatter}
\title{Optimizing UAV-UGV Coalition Operations: A Hybrid Clustering and Multi-Agent Reinforcement Learning Approach for Path Planning in Obstructed Environment}

\author[1]{Shamyo Brotee}
\ead{s.brotee@gmail.com}
\author[1]{Farhan Kabir}
\ead{farhankabir.24csedu.012@gmail.com}
\author[1]{Md. Abdur Razzaque}
\ead{razzaque@du.ac.bd}
\author[2]{Palash Roy}
\ead{palashroy975@gmail.com}
\author[1]{Md. Mamun-Or-Rashid}
\ead{mamun.rashid@gmail.com}
\author[3]{Md. Rafiul Hassan}
\ead{md.hassan@maine.edu}
\author[4]{Mohammad Mehedi Hassan\corref{cor1}}
\ead{mmhassan@ksu.edu.sa}
\cortext[cor1]{Corresponding author}
\address[1]{Green Networking Research Group, Department of Computer Science and Engineering, University of Dhaka, Bangladesh}
\address[2]{Department of Computer Science and Engineering, Green University of Bangladesh, Dhaka, Bangladesh}
\address[3]{College of Arts and Sciences, University of Maine at Presque Isle, ME04769, USA}
\address[4]{Department of Information Systems, College of Computer and Information Sciences, King Saud University, Riyadh, Saudi Arabia}

\begin{abstract}
One of the most critical applications undertaken by coalitions of Unmanned Aerial Vehicles (UAVs) and Unmanned Ground Vehicles (UGVs) is reaching predefined targets by following the most time-efficient routes while avoiding collisions. Unfortunately, UAVs are hampered by limited battery life, and UGVs face challenges in reachability due to obstacles and elevation variations. Existing literature primarily focuses on one-to-one coalitions, which constrains the efficiency of reaching targets. In this work, we introduce a novel approach for a UAV-UGV coalition with a variable number of vehicles, employing a modified mean-shift clustering algorithm to segment targets into multiple zones. Each vehicle utilizes Multi-agent Deep Deterministic Policy Gradient (MADDPG) and Multi-agent Proximal Policy Optimization (MAPPO) — two advanced reinforcement learning algorithms — to form an effective coalition for navigating obstructed environments without collisions. This approach of assigning targets to various circular zones, based on density and range, significantly reduces the time required to reach these targets. Moreover, introducing variability in the number of UAVs and UGVs in a coalition enhances task efficiency by enabling simultaneous multi-target engagement. The results of our experimental evaluation demonstrate that our proposed method substantially surpasses current state-of-the-art techniques, nearly doubling efficiency in terms of target navigation time and task completion rate.
\end{abstract}
\end{frontmatter}
%\maketitle
\section{Introduction}\label{introduction}

The potential of using Unmanned Aerial Vehicles (UAVs) and Unmanned Ground Vehicles (UGVs) in various scenarios has garnered significant attention from researchers in recent years. Regarding revenue, the global autonomous vehicle market for UGVs generated USD 2.24 billion in 2022 and is projected to reach USD 4.08 billion by 2030 \cite{globalMarketUGV}. Simultaneously, the market for UAVs is expected to nearly double in value (from USD 24.9 billion to USD 52.3 billion) over eight years, from 2022 to 2030 \cite{globalMarketUAV}. Currently, these vehicles are employed in diverse domains, including surveillance, search and rescue, inspection, inventory checking, mapping unknown terrains, and planetary exploration \cite{article}. This has led researchers to explore deploying various UAV-UGV combinations for complex tasks like navigation in environments with obstructions. The collaborative UAV-UGV systems have shown promise in intelligence surveillance, reconnaissance, object localization, and crucially, path planning in obstructed environments \cite{articleCaska}.
%%%%%%%%%%%%
%%%%%%%%%%%%
\par UAVs offer the advantage of high altitude and a broad line of sight, while UGVs have longer battery duration and can execute precise operations on ground objects. However, both UAVs and UGVs face significant limitations. UGVs are constrained by limited vertical reach, and UAVs are hampered by a limited power supply, leading to restricted operational range and duration \cite{wang2023air}. Collaborations between UAVs and UGVs can significantly enhance task execution efficiency by overcoming these individual limitations \cite{articleChen}. In collaborative systems, UAVs can access destinations beyond the reach of UGVs. Conversely, UGVs can transport UAVs to conserve their battery life and are better equipped to handle ground objects with greater payload capacity. A homogeneous swarm of UAVs or UGVs often struggles with distant and time-intensive missions, particularly in mission-critical applications. In complex, obstructed environments, a robust UGV can autonomously transport a UAV beyond its flight range, supplying power and substantially increasing its operational range. In turn, the UAV assists the UGV by accessing high-altitude destinations with fewer collision risks. However, integrating multiple UAVs and UGVs for efficient path planning in obstructed terrains poses significant engineering challenges.

Recent research has explored deploying multiple UAVs or UGVs for specific complex tasks. A study proposed a multi-UAV homogeneous system utilizing the Multi-Agent Deep Deterministic Policy Gradient (MADDPG) algorithm for optimized target assignment and collision-free path planning \cite{qie2019joint}. However, this approach does not incorporate UGVs, and many recent works focus on such homogeneous systems. Zhang et al. introduced an Imitation Augmented Deep Reinforcement Learning (IADRL) approach for a UGV-UAV coalition, where both vehicle types complement each other in performing difficult tasks \cite{zhang2020iadrl}. Yet, their methodology does not address UAV-UGV coalitions in terrains with obstacles and lacks diversity in the number of UAVs and UGVs per coalition. These observations have motivated our development of a methodology to enhance collaboration between multiple UAVs and UGVs effectively in obstructed environments.

In this paper, we introduce a novel approach, \textbf{MEANCRFT}, for one-to-many collaboration among UGVs and UAVs, utilizing \textbf{MEAN}-Shift clustering to deploy UAV-UGV \textbf{C}oalitions using \textbf{R}einforcement learning \textbf{F}or reaching aerial and ground \textbf{T}argets. MEANCRFT employs heuristic methods to segment targets into zones, followed by individualized training for vehicles to navigate efficiently in obstructed environments. This approach aims to find fast and collision-free paths toward multiple ground and aerial targets, exploiting the collaborative capabilities of UAVs and UGVs. The environment consists of multiple obstacles, posing a challenge to reach all targets while minimizing time steps and avoiding damage. We employed two Multi-Agent Deep Reinforcement Learning (MADRL) frameworks based on Multi-Agent Deep Deterministic Policy Gradient (MADDPG) and Multi-Agent Proximal Policy Optimization (MAPPO), namely MEANCRFT-MADDPG and MEANCRFT-MAPPO, to train UAV-UGV coalitions for efficient path planning.

Our work introduces new contributions to the field of path planning through the innovative dimension of UAV-UGV coalition. The primary contributions are:
\begin{itemize}
\item Development of a modified mean-shift clustering algorithm that assigns multiple targets to different zones based on target density, deploying multiple UAVs and UGVs within each cluster for complex path-planning challenges.
\item Utilization of two MADRL frameworks, MEANCRFT-MADDPG and MEANCRFT-MAPPO, for training UAV-UGV coalitions in path planning and obstacle avoidance, supporting vehicle count variability in each coalition and enhancing system robustness.
\item Extensive numerical analysis on a custom environment created using OpenAI's gym platform, showing that developed MEANCRFT-MADDPG significantly reduces the number of steps to complete tasks and increases task completion rates compared to state-of-the-art models.
\end{itemize}

This paper is systematically organized: Section \ref{introduction} provides an overview of the UAV-UGV coalition, Section \ref{related_work} reviews relevant literature, Section \ref{system_model} discusses the system model and assumptions, Section \ref{proposed_method} details the proposed approach, Section \ref{experiment} describes the experiment and its results, and Section \ref{conclusion} concludes the paper with a summary and future research directions.

\section{Related Work}\label{related_work}

UAVs and UGVs have seen an unprecedented rise in a multitude of fields, notably path planning, and search and rescue missions. However, due to the limited battery of UAVs and the limited reachability of UGVs, both types of vehicles are restrained in their movement in many path planning and target-reaching scenarios. Many researchers have proposed several approaches to solve this complex path-planning problem for a smart implementation of a functional UAV-UGV coalition.

Bellingham et al. \cite{MUTAPP12} have developed a novel mathematical approach based on Mixed-Integer Linear Programming (MILP) that designs nearly optimal routes toward a goal for multiple UAVs with constraints like no-fly zones and the highest speed and turning rate of the UAVs. Based on genetic algorithms, Shima et al. \cite{MUTAPP17} solved this problem, having special considerations including task prioritization, coordination, time restraints, and flying trajectories. An approach based on particle swarm optimization (PSO) principles was presented by Cruz et al. \cite{PSO}. In recent years, Babel et al. tackled the issue of cooperative flight route planning while minimizing the overall mission time, provided that the UAVs arrive at their destinations simultaneously or sequentially with predetermined time delays \cite{MUTAPP19}.

A computationally effective mathematical model based on cell-winning suppression (CWS) was developed by Jia-lei et al for multi-target task assignment in a complex battlefield environment with attacking and defensive UAVs on two sides \cite{MUTAPP20}. Yan et al. \cite{yan2018path} proposed a path-planning algorithm for the UAVs based on improved Q-learning in an antagonistic environment. A multi-UAV-enabled homogeneous system to optimize target assignment and path planning has been proposed in \cite{qie2019joint}. The authors in \cite{li2020path} have dealt with the path planning problem for a ground target tracking scenario through an obstructed environment by using the line of sight and artificial potential field to construct their reward. Another group of authors in \cite{josef2020deep} introduced a method to implement local online path planning in unknown rough terrain by a UGV. However, the aforementioned works in literature have only considered homogeneous vehicles (Only UAVs or only UGVs) in their respective field. A collaborative approach to heterogeneous vehicles has been ignored in these studies.

%\subsection{MADRL Framework}
To tackle these multiple-agent target assignment and path planning problems, different deep reinforcement learning (DRL) frameworks have been used in the existing literature. Reinforcement learning refers to goal-oriented algorithms that help to attain objectives along a particular dimension over repetitive steps by being rewarded or punished for good and bad actions, respectively \cite{pathmind}.

Usually, a Q-table is maintained to store the state-action pairs in the Q-learning, a basic form of reinforcement learning. When neural networks are associated with reinforcement learning, it is called DRL. Deep Q-Network (DQN), \cite{DQN2015} is an early implementation of DRL, that approximates a state-value function in a Q-Learning framework (Basic reinforcement learning) with a neural network. Consequently, various improvements have come in recent years in the field of DRL. Deep Deterministic Policy Gradient (DDPG), Multi-agent DDPG (MADDPG), Proximal Policy Optimization (PPO), and Multi-agent PPO (MAPPO) are the state-of-the-art additions in this field. DDPG is an actor-critic, model-free algorithm based on the deterministic policy gradient that can operate over continuous action spaces \cite{DDPG}. It combines the actor-critic approach with insights from DQN. MADDPG can be considered just the multi-agent version of DDPG \cite{MADDPG}, and it is considered a state-of-the-art algorithm in the multi-agent reinforcement learning field. On the other hand, PPO is a policy gradient method for reinforcement learning \cite{PPOalgorithms2017}, whereas, MAPPO is essentially the multi-agent version of this \cite{MAPPOmeta2022} \cite{MAPPOsurprising2022} \cite{MAPPOcoordinated2021}.

Qie et al. \cite{qie2019joint} have proposed a multi-UAV-enabled homogeneous system to optimize target assignment and path planning using a MADDPG algorithm, which has achieved excellent performance. However, their developed system is only applicable to homogeneous UAVs. Another group of authors in \cite{10268067} have used a modified K-means clustering algorithm and MAPPO for trajectory design in disaster areas with homogeneous UAVs only. Li et al. \cite{10.3389/fnbot.2022.1076338} have also used PSO-based clustering along with a MAPPO-based algorithm to design a trajectory planning algorithm for situations that require urgent communication. In path planning problems, using a clustering method to manage the targets or destination areas better is a great way to improve performance. Chen et al. \cite{9385941} \cite{9641740} have exploited density-based clustering methods to get improved performances in coverage path planning problems with UAVs. These studies also do not consider UGVs as a coalition approach to solving path-planning problems.

\par Extensive research on coalitions of UAVs and UGVs is a relatively new venture. Ropero et al. designed a UAV-UGV collaborative system for planetary exploration where the goal is to reach a set of target points while minimizing the traveling distance \cite{ropero2019terra}. In a UAV-UGV coalition, UAVs play a major part in planning the path and guiding the collaborative system. Nguyen et al. proposed a solution to a problem where a UAV shepherds a swarm of UGVs \cite{nguyen2020continuous}. Zhang et al. used Imitation Augmented Deep Reinforcement Learning (IADRL), which is a mix of Imitation Learning and DRL, to address the issue of the enormous complexity of UAV-UGV collaboration \cite{zhang2020iadrl}. In their approach, a UGV carries a UAV up to the point where it is unable to navigate vertically to reach the target. The UAV can then be launched to do the assignment and arrive at the destination. In order to increase the operating range of the UAV, the UGV also serves as a charging station. In their work, they assumed their environment without any obstacles and assigned one UAV to a single UGV.

\par Although the above-discussed works have helped improve and overcome a lot of challenges in combined hybrid UAV-UGV systems, there are still some limitations that need to be addressed. Most of the existing works have not considered a collaboration of multiple UAVs and UGVs in an obstructed environment, which enhances the variability of vehicles and the robustness of the system. These observations have motivated us to develop a framework, namely MEANCRFT, to enhance collaboration between multiple UAVs and UGVs in an obstructed environment.

\section{System Model of MEANCRFT}\label{system_model}
\begin{figure*}[!t]
    %\centering
    %\captionsetup{justification=centering}
\begin{subfigure}[H]{.49\textwidth}
    %\centering
    \includegraphics[width=\textwidth]{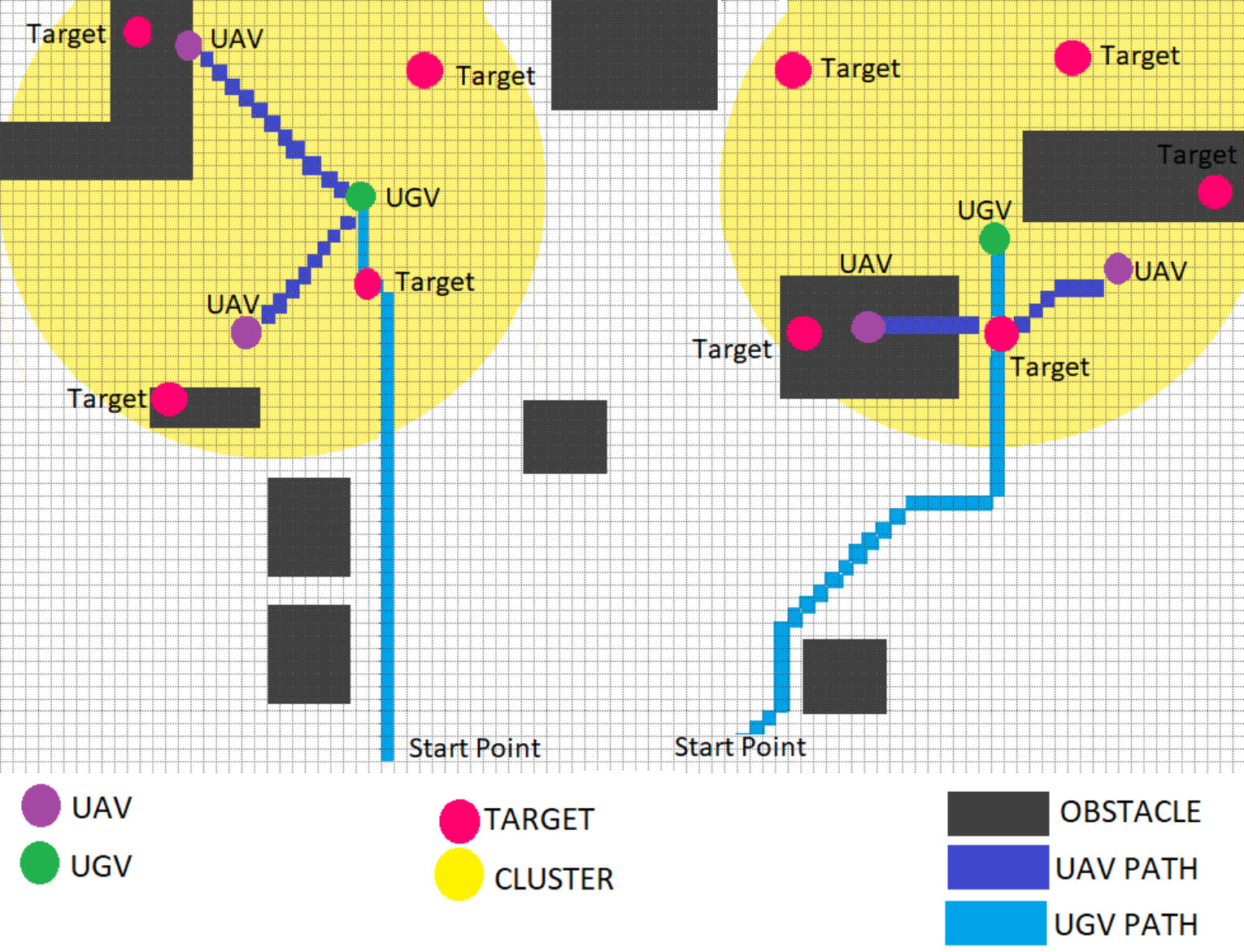}
    \caption{2D grid }~\label{fig:2D Model}
\end{subfigure}
    \hfill
 \begin{subfigure}[H]{.45\textwidth}
    %\centering
\includegraphics[width=\textwidth, height=\textwidth]{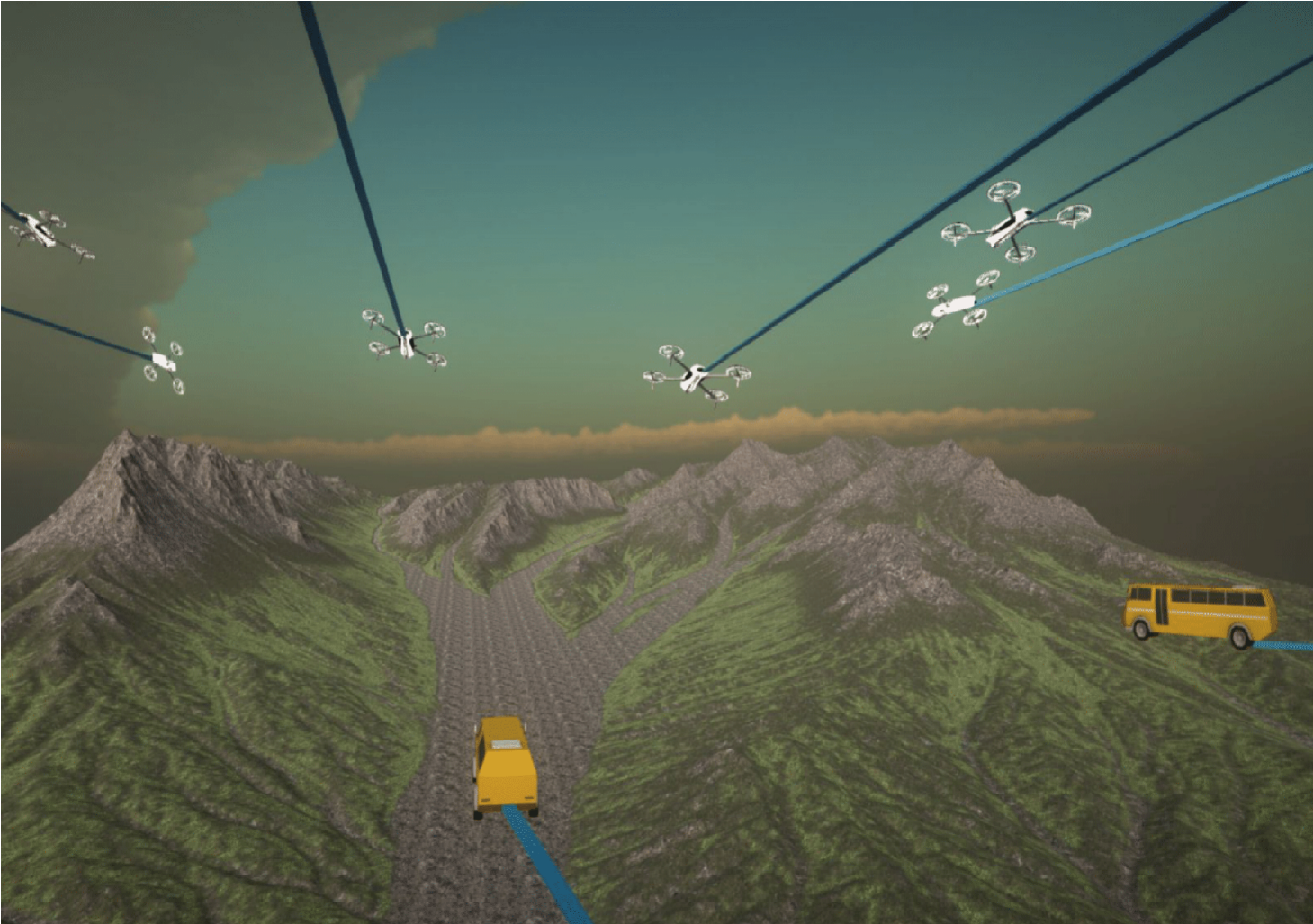}
    \caption{3D representation}~\label{fig:System Model}
\end{subfigure}
\caption{Visual representation of the relationships between key elements}
\end{figure*}

This section provides information about the proposed system along with a detailed depiction of the terms and variables used in it. Fig. \ref{fig:2D Model} depicts the 2D model of different components and their relationship to the developed model.

\begin{table*}[!ht]
\centering
\caption{Notations and their meaning}
\label{tab:notation_table}
\begin{tabular}{l l}
\hline
Notation                  & Description \\
\hline\hline
$\mathcal{A}$ & The set of center points of the unmanned aerial vehicles (UAVs)\\
$\mathcal{G}$ & The set of center points of the unmanned ground vehicles (UGVs)\\
$\mathcal{B}$ & The set of center points of the obstacles\\
$\mathcal{P}$ & The set of center points of the aerial targets\\
$\mathcal{W}$ & The set of center points of the grounded targets\\
$F_a$ & The set of grid points traveled by a UAV $a \in \mathcal{A}$ to reach a target\\
$L_g$ & The set of grid points traveled by a UGV $g \in \mathcal{G}$ to reach a target\\
\begin{comment}
$a_i$ & \textcolor{blue}{action space of $ith$ agent}\\
$o_i$ & observation space of $ith$ agent\\
\end{comment}
$\mu_i$ & deterministic policy of $ith$ agent\\
$\theta$ & Set of policy parameters\\ 
$\nabla_{\theta_i}J(\mu_i)$ & The gradient of $\mu_i$\\
$\mathcal{L}(\theta_i)$ & The loss function\\
$\mathcal{D}$ & Experience Replay\\
\hline\hline
\end{tabular}
\end{table*}

In the proposed 2D environment, UAVs and UGVs navigate to reach the targets by avoiding obstacles, where the set of UAVs and UGVs is denoted by $\mathcal{A}$ and $\mathcal{G}$, respectively. Each aerial target is represented by $p \in \mathcal{P}$ and each grounded target is represented by $w \in \mathcal{W}$, where, $\mathcal{P}$ and $\mathcal{W}$ are the set of the positions of all aerial and grounded targets within a zone. The position of each aerial and grounded target can be denoted by $(x_p, y_p)$, and $(x_w, y_w)$, respectively. The position of a UAV at any time step $t$ is represented by $(x_{a} (t), y_{a}(t))$, where, $a \in \mathcal{A}$ and the position of a UGV at any time step $t$ is 
%$(x_{G}^j (t), y_{G}^j (t))$, where, $j \in \{1,2,\dots, N_G\}$. Here, $N_A$ and $N_G$ denote the number of UAVs and UGVs, respectively. The set of obstacles is denoted by $B$ and each index of an obstacle is represented by $b_k \in B$, where $k \in \{1,2,\dots N_B\}$. Here, $N_B$ is the number of obstacles within a zone. Let $F_{i}$, the path of a UAV in a zone, be 
$(x_{g} (t), y_{g} (t))$, where, $g \in \mathcal{G}$. The set of obstacles is denoted by $B$ and each obstacle position is represented by $(x_{b}, y_{b})$, where, $b \in B$. Let $F_{a} = \{(x_{a} (0), y_{a} (0)), (x_{a} (1), y_{a} (1)),\dots, (x_{a} (n), y_{a} (n))\}$ be the path of a UAV in a zone,
and $L_{g} = \{(x_g (0), y_g (0)), (x_g (1), y_g (1)), \dots, (x_g (n), y_g (n))\}$ be the path of a UGV in a zone, where, $n$ is the number of time steps in an episode.

We assume that UAVs and UGVs are deployed in the environment to reach targets. After reaching a zone, UGVs deploy the UAVs that have been trained to reach the assigned aerial targets while the UGVs get to ground targets. They navigate in the environment avoiding collisions with obstacles and other vehicles. After reaching, the UAVs return to their nearest UGV and the coalitions move to the next zone when the assigned zone has been cleared. In this way, the developed model can cover all the targets receiving minimal damage within a short amount of time. A widespread application of our work can be a post-earthquake rescue mission in a collapsed building, where all the means of ground entry are blocked. In this scenario, the developed model can efficiently plan a path and reach the location of inhabitants who require help. In such an application, high task efficiency and minimal time are of utmost importance, which our model achieves. We have also assumed that the positions of the targets were unchanging. Within a zone, the agents are informed about the position of their reachable targets, the position of the obstacles, and the position of other agents. Moreover, in the 2D environment, the UAVs do not collide with the UGVs as they operate at a higher altitude. Table \ref{tab:notation_table} summarizes the major notations used in this paper.  

\section{Design Details of MEANCRFT}\label{proposed_method}

In this section, we provide a comprehensive explanation of our coalition approach to reach targets in an obstructed environment. 
At first, we will delve into a detailed explanation of the methodology for dividing the targets into multiple circular zones using a heuristic based on a modified version of mean-shift clustering to deploy a specific number of coalitions into those zones for planning a safe route to reach all those targets. In the latter phase, we briefly discuss the two different MADRL models, namely MADDPG and MAPPO, that we used to train our coalitions for efficient path planning and obstacle avoidance.

\subsection{Zone Division: Assignment of Targets into Zones}

The first step of our approach is to divide the targets $P$ into suitable zones. For the selection of zones, we have designed a modified version of the mean-shift clustering algorithm that divides the overall area into circular zones based on target density. Coalitions travel from zone to zone, clearing the targets in them. Algorithm \ref{algoZone} gives a formal depiction of the proposed algorithm for the division and assignment of the targets into zones.
It takes as input a radius value $R$, which is basically the flight range of the UAVs in our system, and the set of points $P$ in the environment. As output, it gives the set of center points of all the circular zones it calculated, $C$.

\subsubsection{Mean-Shift Clustering}
The generic Mean-Shift clustering is a centroid-based sliding window clustering algorithm focused on finding out the center point of each cluster \ref{algo:meanshift}. This algorithm chooses an initial point and creates a sliding window with a specific diameter. This diameter in our model is a function of the UAV flight range. It works by updating candidates for center points to be the mean of the points within the sliding window. 

By moving the center point to the average of the window's points at each iteration, the sliding window is moved toward areas with higher densities. The number of points inside the sliding window determines how dense it is. Naturally, it will eventually gravitate toward locations with larger point densities by adjusting to the window's mean of the points. The sliding window is moved in accordance with the mean of the points inside it and it keeps moving until there is no longer any direction in which a shift can accommodate any center point for the window. This process is continued until all the points fall inside a cluster. As Mean-Shift Clustering is used for identifying clusters, there can be situations where multiple windows overlap. In that case, the sliding window with the most points is kept. The data points are then assigned based on the sliding window that contains them. The detailed step-by-step implementation procedure is described in Algorithm \ref{algo:meanshift}. 
Here, the procedure first accepts as input the radius of the sliding window and the list of data points on which it will perform the clustering. $T$ and $k$ are constant values that are set to accept cluster centers within a specific range.
First, an empty list $U$ is taken which will be used to store the center of the windows. Next, we take a random point as the center of a window and determine the points inside it. We assign these points in a list $Z$ and then determine the centroid of $Z$. This is the current center of the window which is saved in $U$. This process is continued till all the points are inside a window, the center of which is in $U$. $P$ is a list that is used to contain the center of the window in which a point resides. Initially, it is the point itself, and as the process continues, its value is updated (line $20$). This process is repeated at most $k$ times to constantly update $P$ for each point in the given dataset. In each update, the value of the window center can change. The magnitude of this change is determined in line $17$. If this value is more than the threshold value $T$, it means that the window has moved significantly. Otherwise, the shift is not too significant and the current centers are used as the final cluster centers. Finally, after finishing, the non-repeating center values are returned as the cluster centers.
\subsubsection{Modified Mean-Shift Clustering}

\begin{algorithm}[!t]

\caption{Assignment of Targets in Zones}
\label{algoZone}
\begin{algorithmic}[1]
\renewcommand{\algorithmicrequire}{\textbf{Input:}}
\renewcommand{\algorithmicensure}{\textbf{Output:}}
\REQUIRE $R \longleftarrow$\ Radius of UAV Zones, $P \longleftarrow$\ Set of points in the environment
\ENSURE $C \longleftarrow$\ Center of Zones
%$R$ \longleftarrow\ Input Radius \\
%$G$ \longleftarrow\ Empty List \\
\STATE $C \longleftarrow \emptyset$ 
\WHILE{there are points in $P$}
\STATE $G = Mean Shift Clustering (R, P)$
\STATE $C \gets \ C \cup \{g\} , \forall g \in G$
\FOR{each $p \in P$}
\FOR{each $c \in C$}
\IF{$distance(c,p) \le R$}
\STATE $P.delete(p)$
\ENDIF
\ENDFOR
\ENDFOR
\ENDWHILE
\RETURN $C$
\end{algorithmic}
\end{algorithm}
%%%%%%%%%
%%%%%%%%%%%%
\begin{algorithm}[h!]
\caption{Mean-Shift Clustering}
\label{algo:meanshift}
\begin{algorithmic}[1]
\renewcommand{\algorithmicrequire}{\textbf{Input:}}
\renewcommand{\algorithmicensure}{\textbf{Output:}}
%\PROCEDURE{Modified Mean_Shift Clustering}{($R, P$)}
\PROCEDURE{MeanShiftClustering}{$R, P$}
%\REQUIRE $R \longleftarrow$\ Radius of UAV Zones, $P \longleftarrow$\ Set of points in the environment
%\ENSURE $O \longleftarrow$\ Center of Clusters
\STATE $T \gets Constant$
\STATE $k \gets Constant$
\WHILE{$k$}
\STATE $U \longleftarrow \emptyset$ 
\FOR{$p \in P$}
\STATE $Z \longleftarrow \emptyset$
\FOR{$p' \in P$}
\IF{$distance(p, p') \leq R$}
%\STATE $Z.append(P[j])$
\STATE $Z \gets Z \cup p'$
\ENDIF
\ENDFOR
\STATE $V = mean(Z)$
\STATE $U \gets U \cup V$
\ENDFOR
\STATE $M = mean(U - P)$
\IF{ $M \le T$}
\STATE \textbf{break}
\ENDIF
\STATE $P = U$
\STATE $k \gets k + 1$
\ENDWHILE
\STATE Initialize $O$ as empty set
\STATE $O \gets O \cup p,$ $\forall p \in P$
\RETURN $O$
\ENDPROCEDURE
\end{algorithmic}
\end{algorithm}

The developed Algorithm \ref{algoZone} is based on a modified version of the aforementioned Mean-Shift clustering algorithm. Firstly, we take the radius of the zones $R$ and the list of the targets' coordinates $P$ in the environment as inputs. The radius of the zones is a preset value that is dependent on the functional range of the drones. Next, in line $3$, we use Mean-Shift clustering of Algorithm \ref{algo:meanshift} which returns the centers of the densest zones in the environment that can form a cluster. We add them to a list $C$ and use it to find out all the coordinates that are inside a zone by comparing their distances with the center points stored in $C$. Then we delete the points from the coordinate list $P$ that are within $R$ distance from each element in $C$ since they are already assigned to a zone. We keep repeating this process of calling the Mean-Shift clustering algorithm and assigning the points into zones until all the points in $P$ are assigned properly into a zone. This returns new center points and the process continues until no points are left unassigned and all the center points are in the list $C$. This is how our proposed heuristic works, based on a densest-area-first approach. 
%It chooses a random point as the center $O$ of each circle. If the mean position of points $C$ inside the circle changes, it moves the circle's center $O$ to the new mean position and \RE{recalculates the mean $C$}. If the mean position stays the same, it adds the center of the circle $O$ to a group $G$ which is our list of zones, \RE{removes the points from the environment} and continues the process all over again with a new random point until all the points are assigned to a zone. Finally, it returns $G$ which is the list of all zones.

\par We have designed this zoning algorithm because we want to use the battery capacity and flight range of the UAV as decisive factors for choosing our zones. As every zone has to be covered by the coalitions to reach all available targets in the environment, dividing the targets based on density reduces the overall zone count. Again, as the number of UAVs in a coalition can be variable, we ensure maximum target-reaching efficiency within a zone by sending maximum available agents in it to reduce overall time.

After dividing the target areas into zones, one or multiple coalitions are sent to each zone based on the requirements. When a coalition enters a zone, the UAVs can be deployed to reach the aerial targets by avoiding obstacles and other agents. In the meantime, the UGV in the coalition can navigate through the obstructed terrains and reach the ground targets. When the battery of a UAV reaches a critical level, the target assignment for the UAV will stop and it will immediately land on the nearest available UGV. This ensures maximum utilization of the battery life of the UAV and also clears the zones rapidly.

\subsection{Assignment of Coalition to Zones}
\begin{figure}[!t]
\centering
\includegraphics[width=.95\columnwidth,keepaspectratio]{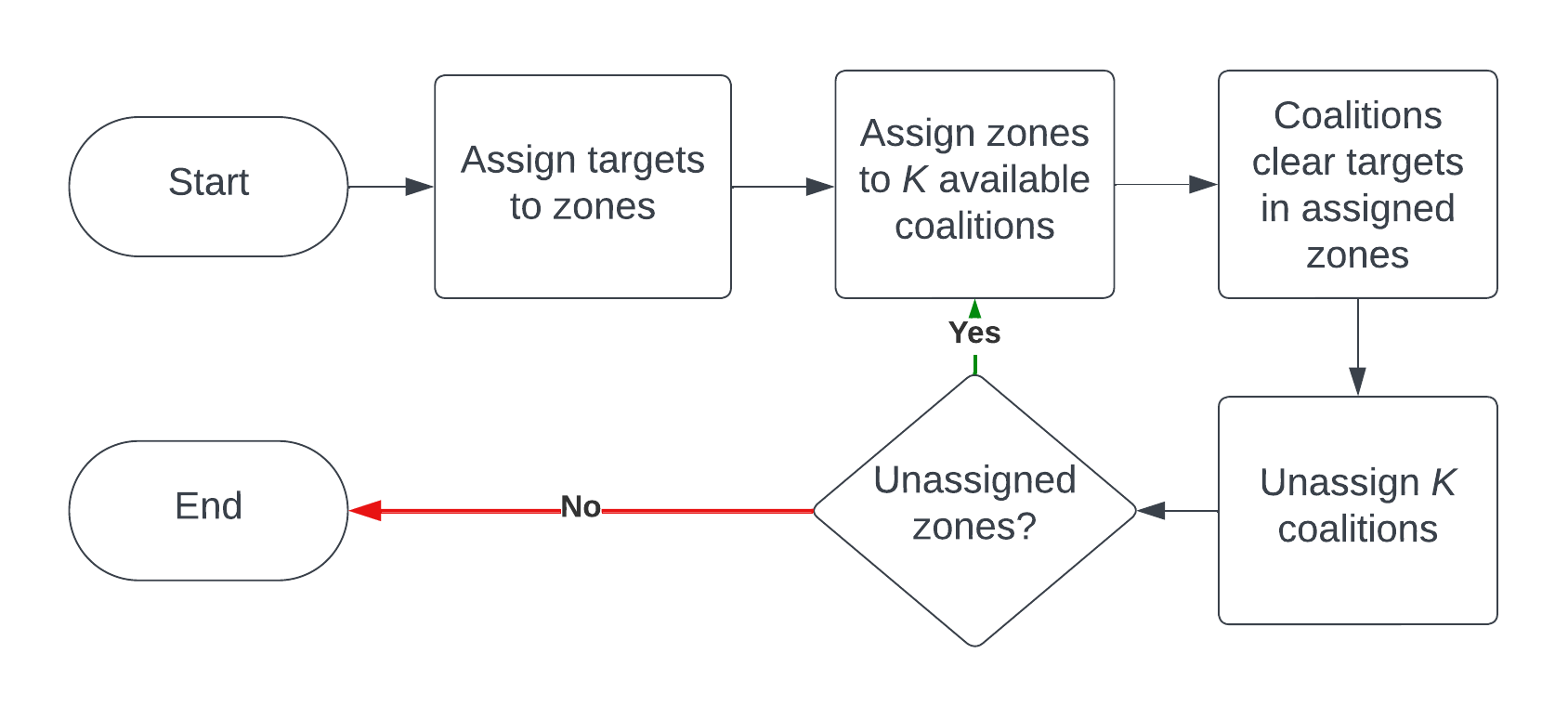}
\caption{Flowchart of the proposed solution framework}~\label{fig:SolutionFramework}
\end{figure}

After grouping all the targets into various circular zones, we deploy MADRL-trained coalitions in each of those zones to reach the targets while avoiding collisions. The sequential steps for our solution to assign the coalitions into zones are depicted in Fig. \ref{fig:SolutionFramework} and the step-by-step procedure can be described as follows,
\begin{enumerate}
        \item At first, the number of coalitions, $K$ are sent to a zone.
    \item After that, the trained models of UAVs and UGVs are deployed to cover the targets. A coalition is considered available only when all the deployed UAVs have landed on the available UGVs after clearing the zone.
    \item When a zone is cleared, the previously assigned $K$ coalitions are available, which are then assigned to the next target zone.
    \item Steps 2 and 3 are repeated until all the zones have been cleared, and subsequently all the targets are reached.
\end{enumerate}
%%%%%%%%%%%%
%%%%%%%%%

The coalitions will follow a specific set of steps when they reach the zone that they are assigned to: the UGVs will navigate to reach the ground targets after deploying UAVs. The UAVs will also roam to reach the aerial targets. During this process, if a UAV needs to land, it will signal its nearest UGV. The UGV will pause its current activity, wait for the UAV to land, and after landing, resume its function. 

It should be noted that we have trained the UGVs first as they require less complex inputs for training. This is because the only function a UGV has is to reach its assigned target. For actions such as UAV landing, it only needs to pause its function which does not require training. On the other hand, UAVs need to reach their assigned targets and return to their nearest UGV. This makes the task more intricate and also requires the information of the nearest UGV in their observation space. This is why, we first train the UGVs and use this trained model to get some UGV coordinates in an environment. Next, in the UAV training scenario, we used these demo coordinates as input in the observation space of the UAV. This way, the first half of the UAV training is to reach the assigned target. The next half is to "land" on the nearest UGV, the position of which is provided by our demo data. In this way, we divided our training into two parts and used our UGV-trained model to train our UAVs. Ultimately, these two separate training models are combined in the full simulation environment. 

There, the available targets are first divided into zones. Then only the trained models for UGVs are used to move coalitions to their assigned zones. After reaching the zones, the UGVs deploy UAVs and together they move in the environment to reach all the targets inside the zone. The UGVs will navigate in the environment to reach their assigned targets avoiding obstacles and other UGVs, while at the same time, UAVs will also move around the environment avoiding obstacles and other UAVs to reach assigned aerial targets. When UAVs are done reaching their respective targets, they move to their nearest UGV location. The selected UGV for landing pauses its model until the UAV lands. After that, it resumes its model to reach ground targets.

\subsection{Constraints for MADRL Training}

From Table \ref{tab:notation_table}, we know that $F_a$ and $L_g$ are the path lengths of UAV $a \in \mathcal{A}$ and UGV $g \in \mathcal{G}$, respectively. The goal of this work is to minimize these path lengths while keeping the journey of the vehicle safe and free from collisions, which can be denoted as,

\begin{equation}
   min (\sum _{a \in \mathcal{A}}  F_{a}
+
    \sum _{g \in \mathcal{G}}  L_{g})
\end{equation}

There are certain constraints that we need to follow while achieving the sole objective. The following constraints are set that need to be satisfied:

    \begin{equation}\label{UGV-target}
        (x_p, y_p) \in F_a, \quad \forall p \in \mathcal{P}, \quad \exists a \in \mathcal{A}
    \end{equation}
    
    \begin{equation}\label{UAV-target}
          (x_w, y_w) \in L_g, \quad \forall w \in \mathcal{W}, \quad \exists g \in \mathcal{G}   
    \end{equation}
    
    \begin{equation}\label{UGV-path}
           (x_{b}, y_{b}) \notin F_a, \quad \forall b \in \mathcal{B}, \quad \forall a \in \mathcal{A}
    \end{equation}
    
    \begin{equation}\label{UAV-path}
            (x_{b}, y_{b}) \notin L_g, \quad \forall b \in \mathcal{B}, \quad \forall g \in \mathcal{G}
    \end{equation}

    \begin{equation}\label{UAV-coll}
            (x_a(t), y_a(t)) \not= (x_\textsl{a'}(t), x_{{a'}}(t)), \quad \forall a,{a'} \in \mathcal{A}
    \end{equation}
    
    \begin{equation}\label{UGV-coll}
            (x_g(t), y_g(t)) \not= (x_{g'}(t), x_{{g'}}(t)), \quad \forall g, g' \in \mathcal{G}
    \end{equation}

    \begin{equation}\label{UAV-return}
        \forall a, (x_a(n), y_a(n)) = (x_g(n), y_g(n)) 
    \end{equation}

At a time, each UAV and UGV is assigned one aerial target and one grounded target, respectively. Constraint in Eq. \eqref{UGV-target} and \eqref{UAV-target} specify that every aerial target is covered by a UAV and every ground target is covered by a UGV. Moreover, the path of a UGV and a UAV is free from collisions with any obstacle, as denoted in constraint \eqref{UGV-path} and \eqref{UAV-path}. Constraint (\ref{UAV-coll}) and (\ref{UGV-coll}) denote that each UGV $g \in \mathcal{G}$ must not collide with another UGV $g' \in \mathcal{G}$ and each UAV $a \in \mathcal{A}$ must not collide with another UAV $a' \in \mathcal{A}$, respectively. Constraint in Eq.(\ref{UAV-return}) specifies that a UAV must come back to its nearest UGV and have the same position as the UGV at the $n$-th step, which is the final step of an episode.

\subsubsection{MADDPG Framework}
Multi-Agent Deep Deterministic Policy Gradient (MADDPG) algorithm is a very well-known technique for path planning or localization tasks in recent eras, which has grown to be a state-of-the-art strategy to solve a multitude of such problems. In the proposed method, multiple UGVs and UAVs are deployed to reach targets in a rough obstructed environment. These vehicles cooperate with each other to reach targets spread all over the environment. The previous section described the detailed methodology to divide all the targets into zones and assign coalitions to go to each zone based on the flight range of the UAVs. In this section, we describe how we exploit one of the MADRL strategies, namely MADDPG, to train the coalitions to efficiently reach the targets while avoiding obstacles. In the proposed training approach, the system is divided into 2 separate environments where the UGVs and UAVs are trained separately, and then both of the trained models are evaluated in a combined environment where they are deployed together to work cooperatively.
\par MADDPG solves the target assignment and path planning problem by modifying Deep Deterministic Policy Gradient (DDPG) \cite{MADDPG} and is considered a multi-agent variation of DDPG. Its core concept is to centralize training while decentralizing execution (CTDE). Because they do not utilize the information of other agents, the deep Q-learning network (DQN) and DDPG perform poorly in multi-agent systems. The MADDPG method gets around this problem by relying on the observations and actions of other agents. There are two networks for each agent: an actor network and a critic network. The actor network calculates the action to be taken based on the agent's current state, whereas the critic network assesses the action taken by the actor network in order to improve the actor network's performance. An experience replay buffer is employed to retain a set amount of training experience and random reads are performed on it to update the network. This is done to break correlations in the training data and avoid divergence. The actor network only gets observation information during the training phase, however, the critic network gets information from other agents' actions and observations. The critic network is only involved in the training phase, and after training, every agent uses only the actor network. As the actors are not connected to each other and guide the agents individually in the execution phase, decentralized execution is achieved.

\par In our system, we consider each of the vehicles to be an agent. The agents in each of the training phases have an action space and a state space. The action space of each UGV and UAV consists of its velocities along both the $X$ and $Y$ axes in a geometric system. Therefore, the action space of UAV $a \in \mathcal{A}$ can be defined as $\xi_{a} = ({u_a, v_a})$; and the action space of UGV $g \in \mathcal{G}$ as $\xi_{g} = ({u_g, v_g})$. The state space of each UGV and each UAV is denoted by $o_{g}$ and $o_{a}$, respectively. The state space of a specific UGV agent $o_{g}$ consists of the relative position of all the grounded targets in the environment, the relative position of the obstacles in the environment, the agent's own velocity as well as its own position, and the relative position of other UGVs in respect to the agent. Similarly, in the case of UAVs, the state space of a specific UAV agent $o_a$ consists of the relative position of all the targets and obstacles, the relative positions of other UAVs with respect to the agent, and the agent's own velocity and position. However, as UAVs require a position to land after reaching targets; therefore, the relative position of its nearest UGV is also added, along with a binary variable that indicates whether the UAV has reached its assigned target or not. This value is 0 when the target has not been reached and it is 1 when the UAV reaches the target. This informs the UAV that it has reached its assigned target and now it can start the process of landing on a UGV. Both of these are additions to a UAV agent's state space. These inputs are used to train the models in separate environments. After the training phases are done, the trained models are deployed in a combined environment where both the UGVs and UAVs collaborate to reach the targets in the obstructed environment. 

In each training scenario, there are $N$ agents. Each agent has its own deterministic policy $\mu_i$. The policies are parameterized by $\theta = \{\theta_1, \theta_2, \dots , \theta_N\}$. For an agent $i$, the gradient of the deterministic policy $\mu_i$ can be written as
\begin{equation}
\label{eq:gradMADDPG}
    \nabla_{\theta_i}J(\mu_i) = \mathbb{E}_{x,a\sim \mathcal{D}}[\nabla_{\theta_i}\mu_i(a_i|o_i)\nabla_{a_i}Q_i^\mu(x,a_i,a_2,\dots , a_N)|_{a_i = \mu_i(o_i)}],
\end{equation}
where, $x$ is the state space which is equal to $\{o_1, o_2, \dots , o_N\}$, $Q_I^\mu(x, a_1, a_2, \dots, a_N)|_{a_i = \mu_i(o_i)}$ is the $Q-value$ function, $a_i$ is the action of the $i-th$ agent $o_i$ is the observation of each $ith$ agent, $\mathcal{D}$ is the experience replay where each tuple is $(x,x',a_1,\dots,a_N,r_1,\dots,r_N)$ as a record of the experiences of all agents. After the actions of all agents have been executed, $x'$ is used to represent the new state of the environment. $r_i$ is the reward of the $ith$ agent. The loss function, $\mathcal{L}(\theta_i)$ updates the critic network, $Q_i^\mu$ which is given by
\begin{equation}
\label{eq:lossMDDPG}
    \mathcal{L}(\theta_i) = \mathbb{E}_{x,a,r,x'}[(Q_i^\mu (x,a_i,a_2,\dots,a_N) - y)^2],
\end{equation}
where,
\begin{equation}
    y = r_i + \gamma Q_i^{\mu'}(x',a_i',\dots,a_N')|_{a_j' = \mu_j'(o_j)}.
\end{equation}

\begin{algorithm}[!t]

\caption{MADDPG pseudo-code}
\label{algoMADDPG}

\begin{algorithmic}[1]
\STATE $E \longleftarrow$\ Number of episodes 
\STATE $T\longleftarrow$\ Number of time steps 
\STATE $N\longleftarrow$\ Number of agents 
\STATE Initialize replay buffer $\mathcal{D}$
\FOR{$Episode = 1 $ to $E$}
    \STATE Initialize a random process $\mathcal{M}$ for action exploration
    \STATE Receive initial state $x$
    \FOR{$t = 0 $ to $T$}
        \FOR {$i=1$ to $N$}
            \STATE select action \(a_i = \mu_{\theta}(o) + \mathcal{M}_t\)
        \ENDFOR
    \ENDFOR
    \STATE Execute actions $a = (a_{1},\dots,a_{N})$ and observe reward $r$ and new state $x'$ 
    \STATE Store $(x,a,r,x')$ in replay buffer $\mathcal{D}$
    \FOR{$i=1$ to $N$}
        \STATE Sample a random minibatch of $S$ samples $(x, a, r, x\textsuperscript{'})$ from $\mathcal{D}$
        \STATE Set $y = r_i+\gamma Q_i^{\mu'}({x'}, a_1',\dots, a_N')|_{a_k' = \mu_k'(q_k)}$
        \STATE Update critic of agent $i$ by minimizing the loss using Eq. \eqref{eq:lossMDDPG}
        %\begin{equation}
        %\(\mathcal{L}(\theta_i) = \frac{1}{S}{\sum}_j(y^i - Q_i^{\mu}(x^j,a_i^j,\dots,a_N^j))^2\)
        %\end{equation}
        \STATE Update actor of agent $i$ using the sampled policy gradient via Eq. \eqref{eq:gradMADDPG}
        %\begin{equation}
        %\(\nabla_{\theta_i}J(\mu_i)\approx \frac{1}{S}\sum_j\nabla_{\theta_i}\mu_i(o_i^j)\nabla_{a_i}Q_i^\mu(x^j,a_1^j,\dots,a_N^j)|_{a_i=\mu_i()o_i^j}\)
        %\end{equation}
    \ENDFOR
    \STATE Update target network parameters for agent $i$:\\
    \centering{\(\theta_i' \xleftarrow{}\tau\theta_i + (1-\tau)\theta_i'\)}
\ENDFOR
\end{algorithmic}
\end{algorithm}

\subsubsection{MADDPG Reward Calculation} \label{rewardcalc}

For training the vehicles in their respective environments, a multi-agent deep reinforcement learning setup has been exploited. In such setups, appropriate reward structures are necessary in order to get acceptable behavior from agents, i.e., the UAVs and UGVs in their respective environments. In the developed UAV and UGV environments, multiple homogeneous agents are trained to reach the available targets in an acceptable manner. To attain such outputs, we have to come up with a reward structure that encourages multiple agents to reach their targets efficiently. At the same time, the agents also need to avoid any obstacle collisions while on their way to the targets. For this, the assigned reward value for each agent is divided into some components which are affected by some parameters of the environment and are different for each environment type. For encouraging the agents to reach targets, the distance between agents and targets is considered. For collision avoidance, the agent-agent distance as well as the agent-obstacle distance is also considered. It should be noted that for various reward components, a function $T_i(.)$ is used which accepts distance as input and returns a reward value.
\begin{enumerate}

    \item \textit{UGV Reward Components}
    \begin{enumerate}
        \item {Reward Based on Distance to Targets:} In each type of environment, there can be multiple targets with multiple agents where the agents together need to find a way to reach all the available targets in an efficient manner. To do so, a reward component $r_1$ is assigned based on the distance between all targets and agents. Specifically, the minimum distance between available targets and agents is taken into consideration. To assign such a reward, at first, the distance between all UGV agents to each ground target $w$ is calculated and stored in a set $d_w$. From all these values, the minimum distance, $\min(d_w)$ of each ground target $w$ to its closest UGV is taken. After calculating, we sum the minimum values and assign this negative reward component to each agent. 

\begin{equation}
    r_1 = r_1 + T_1( - min(d_w)), \quad \forall w \in \mathcal{W}
\end{equation}

\begin{figure*}[!t]
    \centering
    \captionsetup{justification=centering}
\begin{subfigure}[H]{0.241\textwidth}
    \centering
    \includegraphics[width=\textwidth]{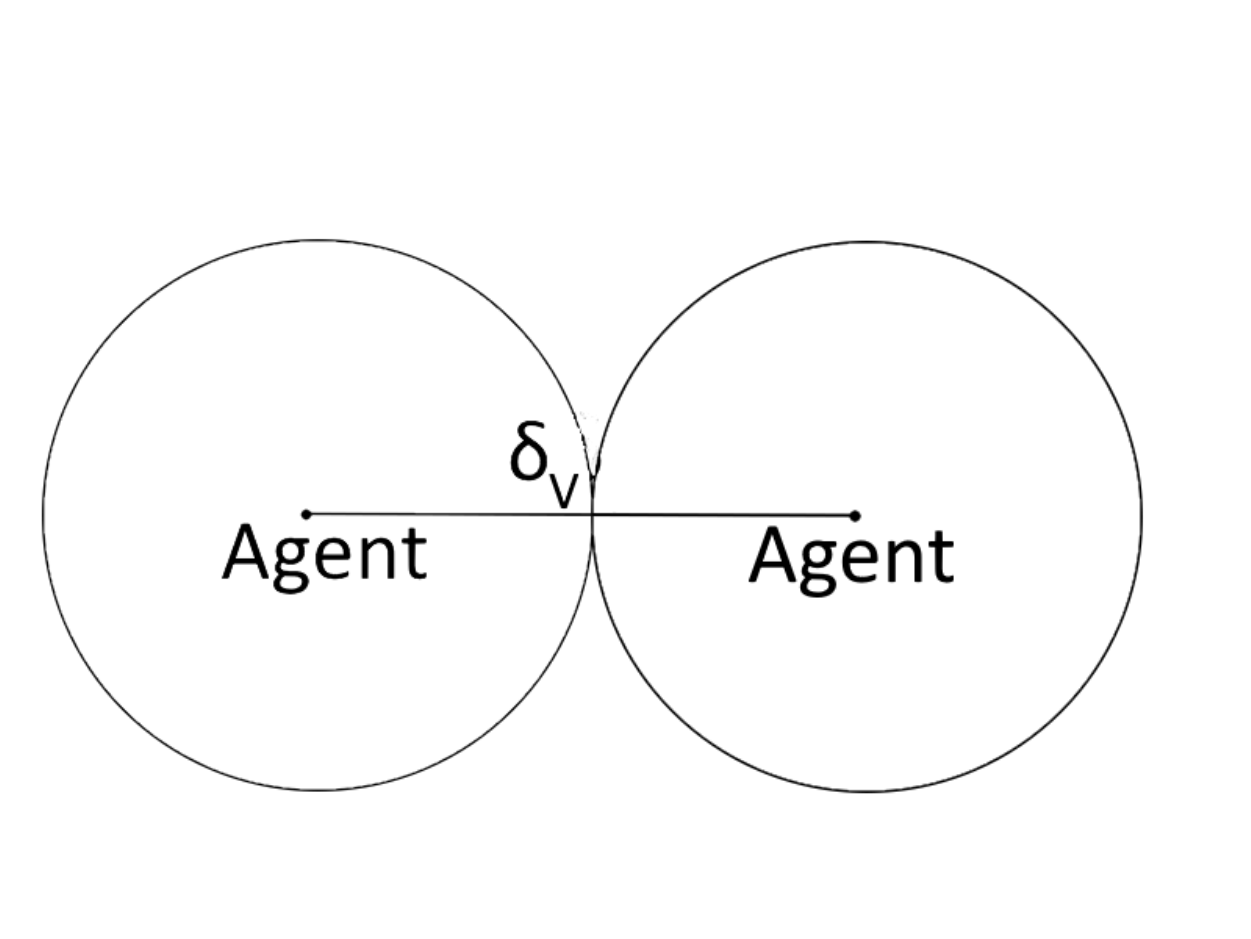}
    \caption{$\delta_V$}
    \label{fig:1a}
\end{subfigure}
    %\hfill
 \begin{subfigure}[H]{.2\textwidth}
    \centering
    \includegraphics[width=\textwidth]{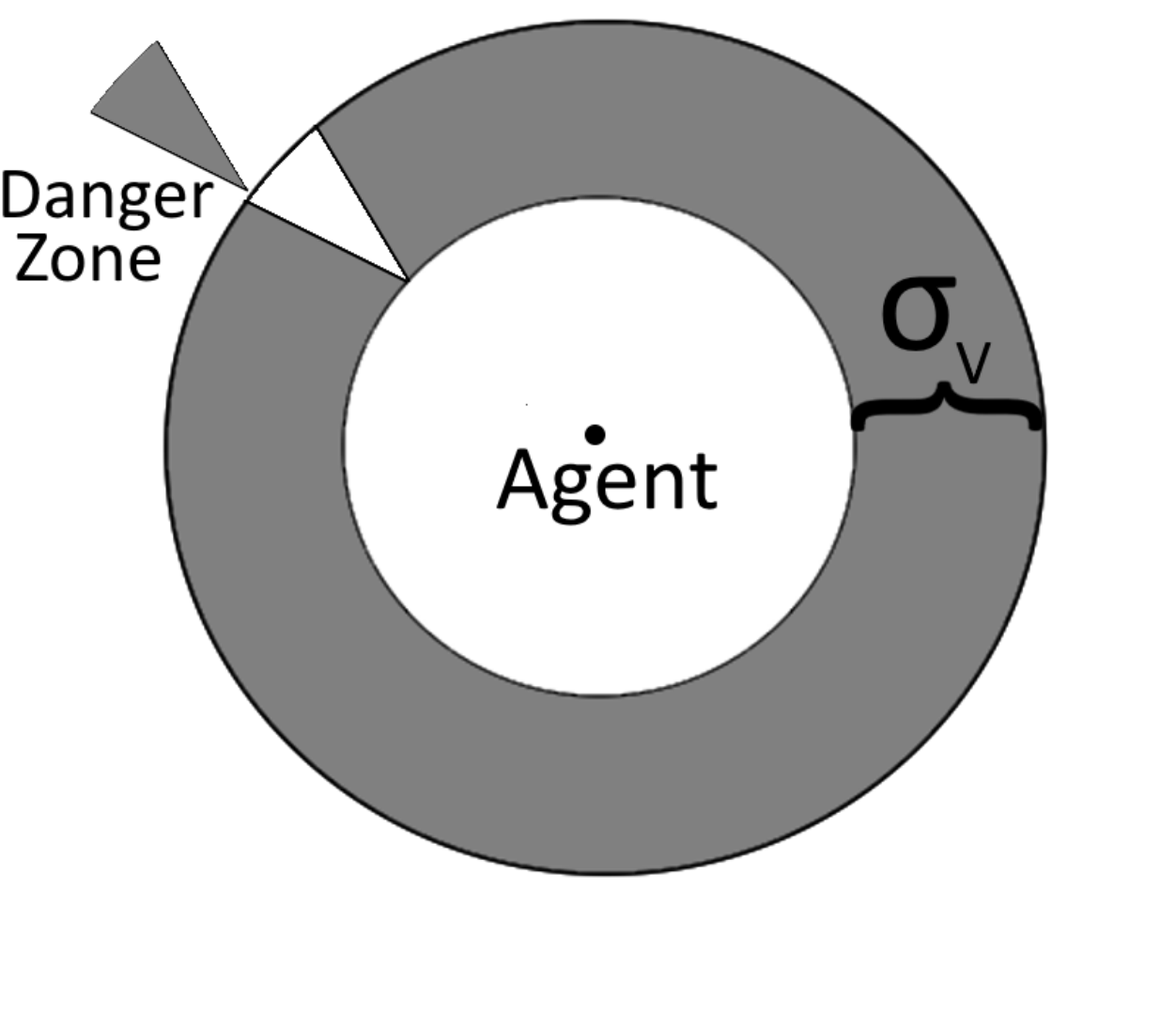}
    \caption{$\sigma_V$}
   \label{fig:1b}
\end{subfigure}
%\hfill
\begin{subfigure}[H]{.241\textwidth}
    \centering
\includegraphics[width=\textwidth]{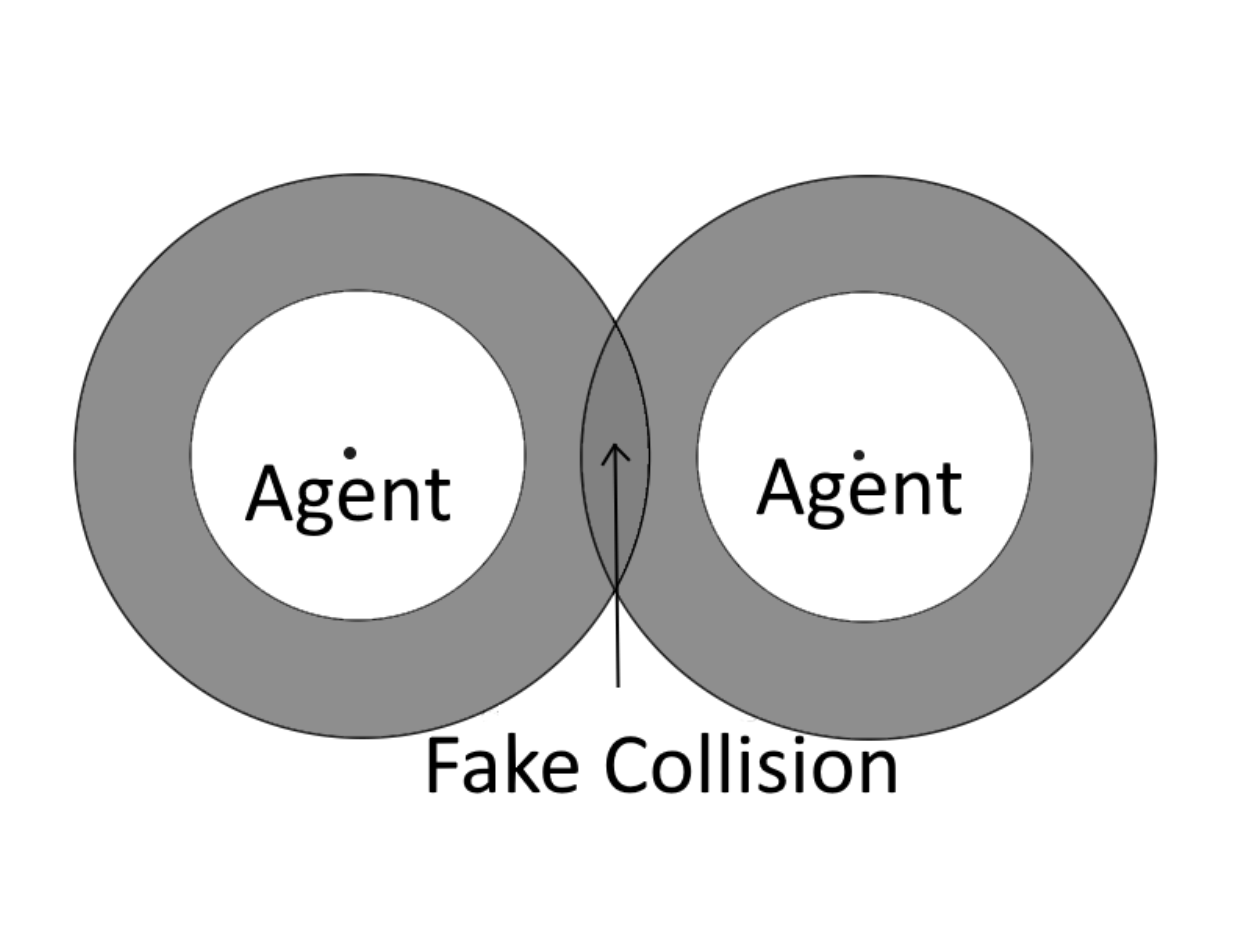}
     \caption{Fake collision}
   \label{fig:1c}
\end{subfigure}
\begin{subfigure}[H]{.241\textwidth}
    \centering
\includegraphics[width=\textwidth]{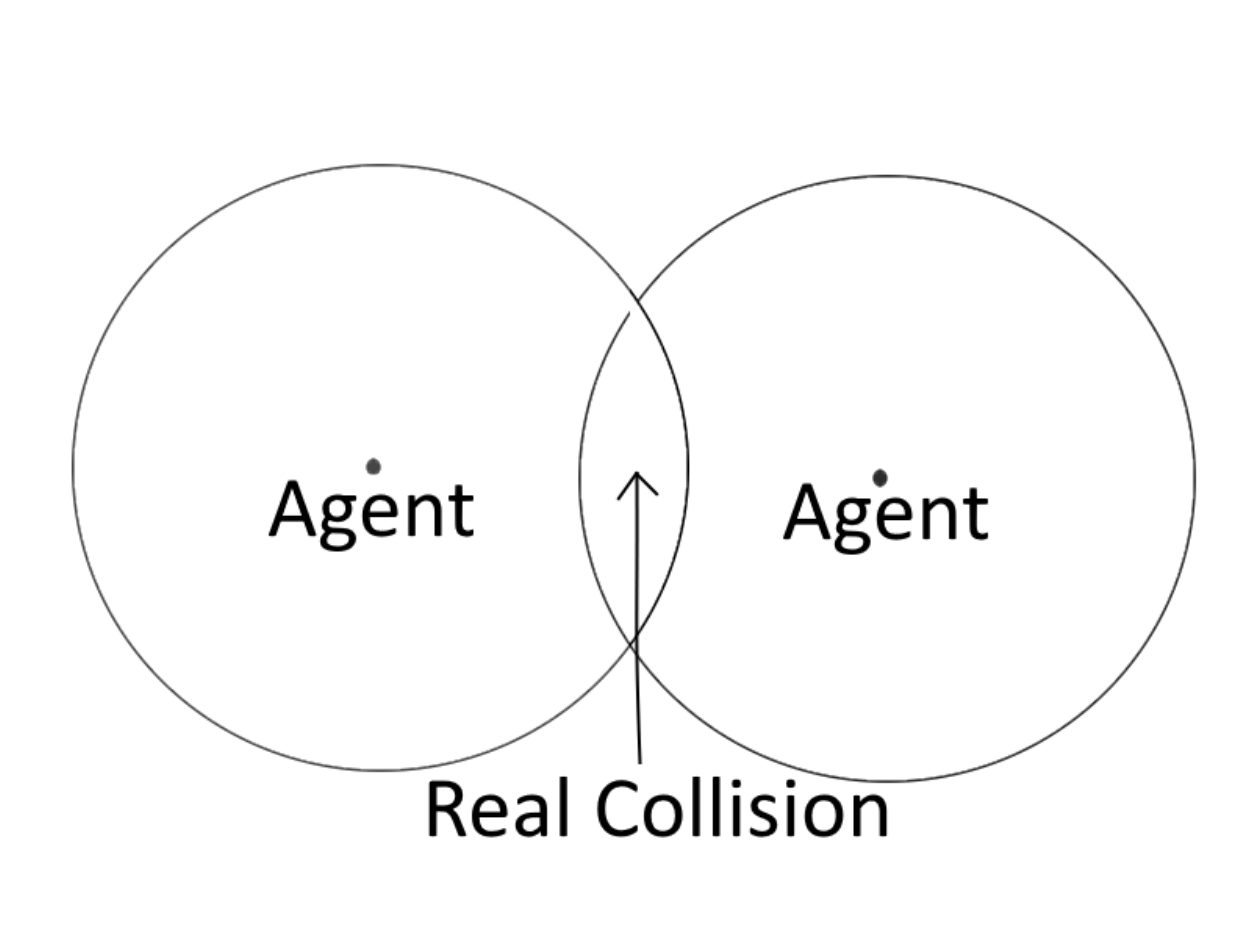}
     \caption{Real collision}
   \label{fig:1d}
\end{subfigure}
\caption{Agent-agent collision}
 \label{ag-ag_col}
\end{figure*}

        \item {Reward Based on Agent-Agent Collision:} While navigating the environment, the agents must avoid colliding with other agents. For this, a reward component, $r_2$ has been derived, where the agents are punished for being within collision distance of each other.

Here, we have considered an area around each agent and named it \textbf{Danger Zone}, whose width is \(\sigma_V\) and \(\delta_V\) is the maximum distance between agents during which collision happens. Fig. \ref{ag-ag_col} shows an example scenario of the danger zones for agents. Whenever the danger zones of two agents overlap or collide, i.e., the distance between agents, $d_V < (\delta_V + \sigma_V)$, we call it fake collision. A negative reward is given to the fake colliding agents based on the size of the overlap area. The more the overlapping area, the more the magnitude of the negative reward. If agents actually collide, the agents get the maximum negative reward value and the collision information is recorded for evaluation. The reward value can be estimated as follows,
\begin{equation}
    r_2 = r_2 - T_2(\delta_V + \sigma_V - d_V)
\end{equation}

\begin{figure*}[!t]
    \centering
    \captionsetup{justification=centering}
\begin{subfigure}[H]{0.241\textwidth}
    \centering
    \includegraphics[width=\textwidth]{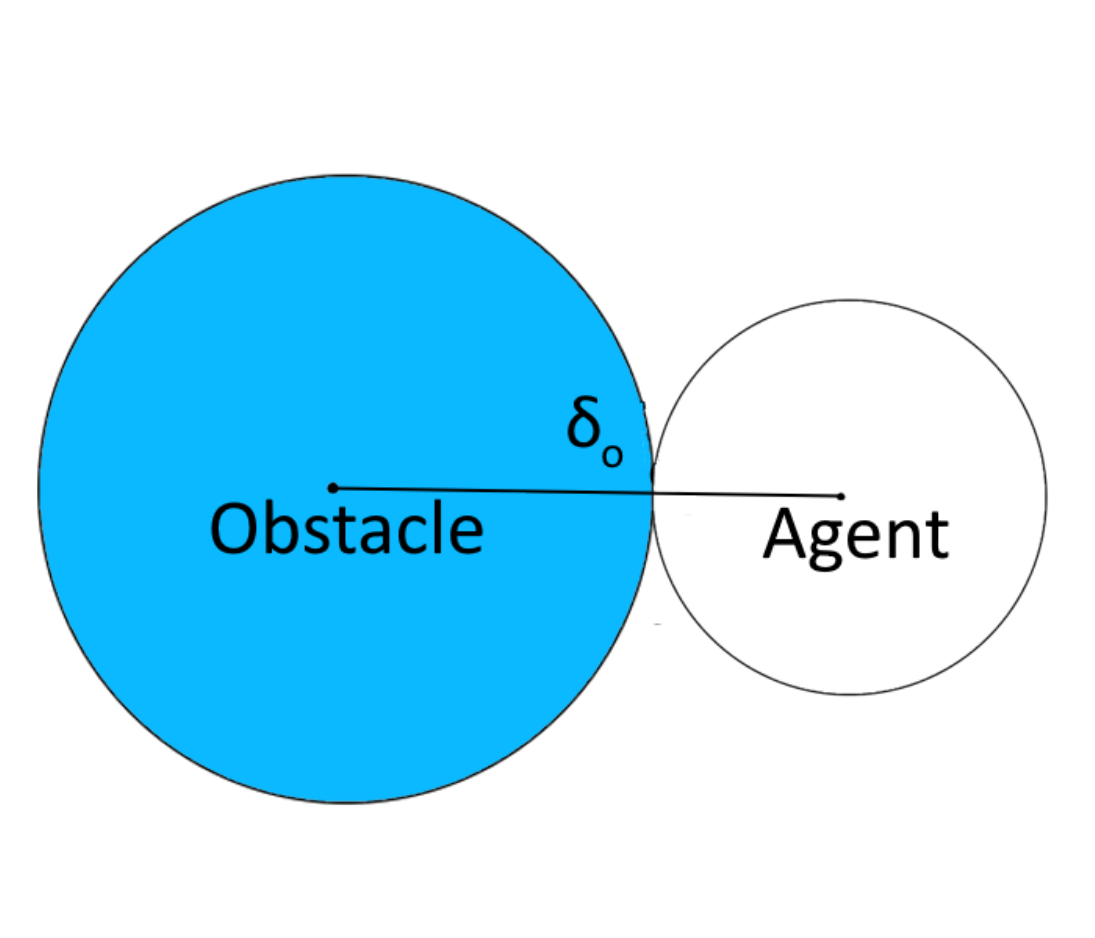}
    \caption{$\delta_O$}
    \label{fig:2a}
\end{subfigure}
    %\hfill
 \begin{subfigure}[H]{.2\textwidth}
    \centering
    \includegraphics[width=\textwidth]{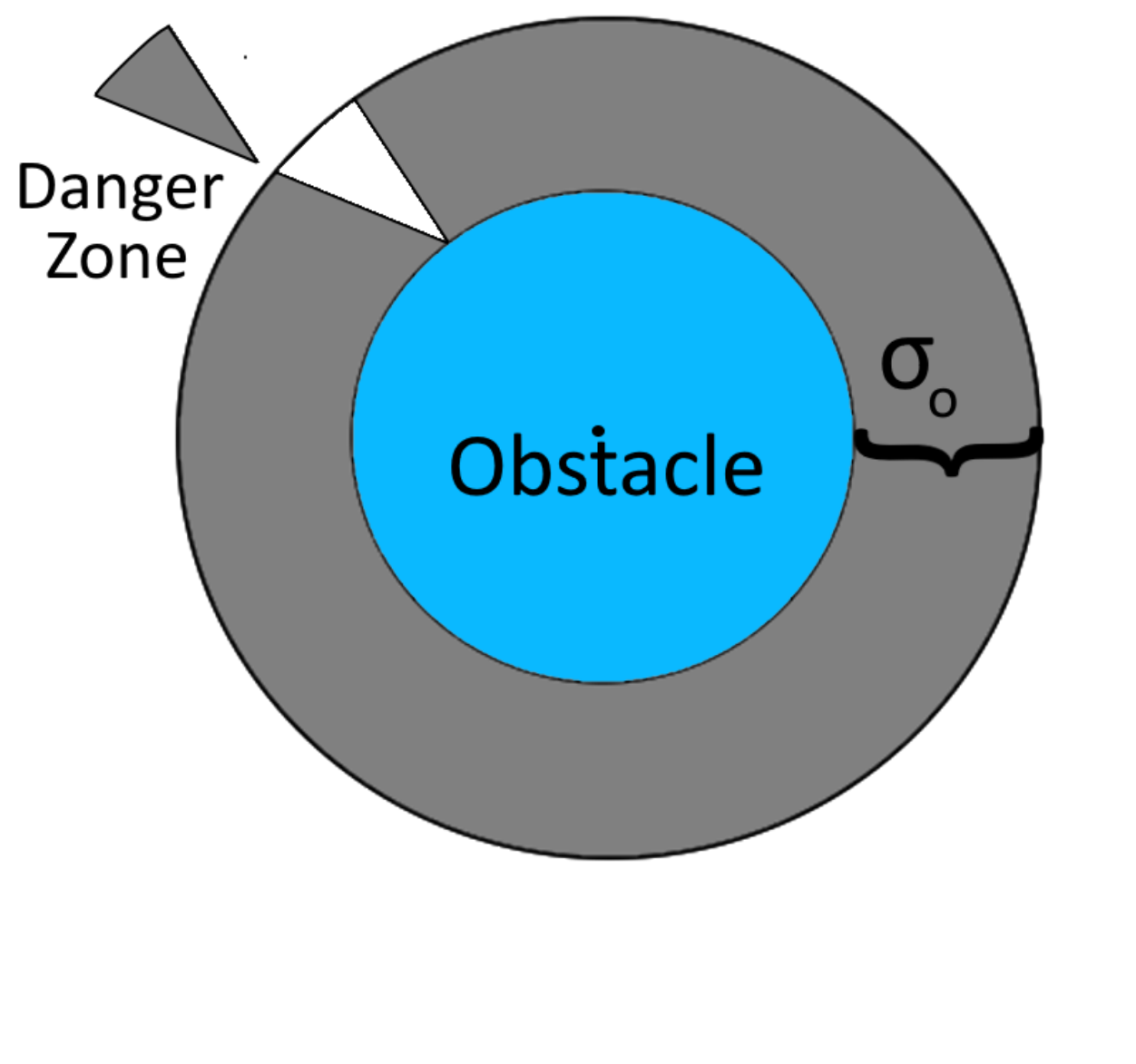}
    \caption{$\sigma_O$}
   \label{fig:2b}
\end{subfigure}
%\hfill
\begin{subfigure}[H]{.241\textwidth}
    \centering
\includegraphics[width=\textwidth]{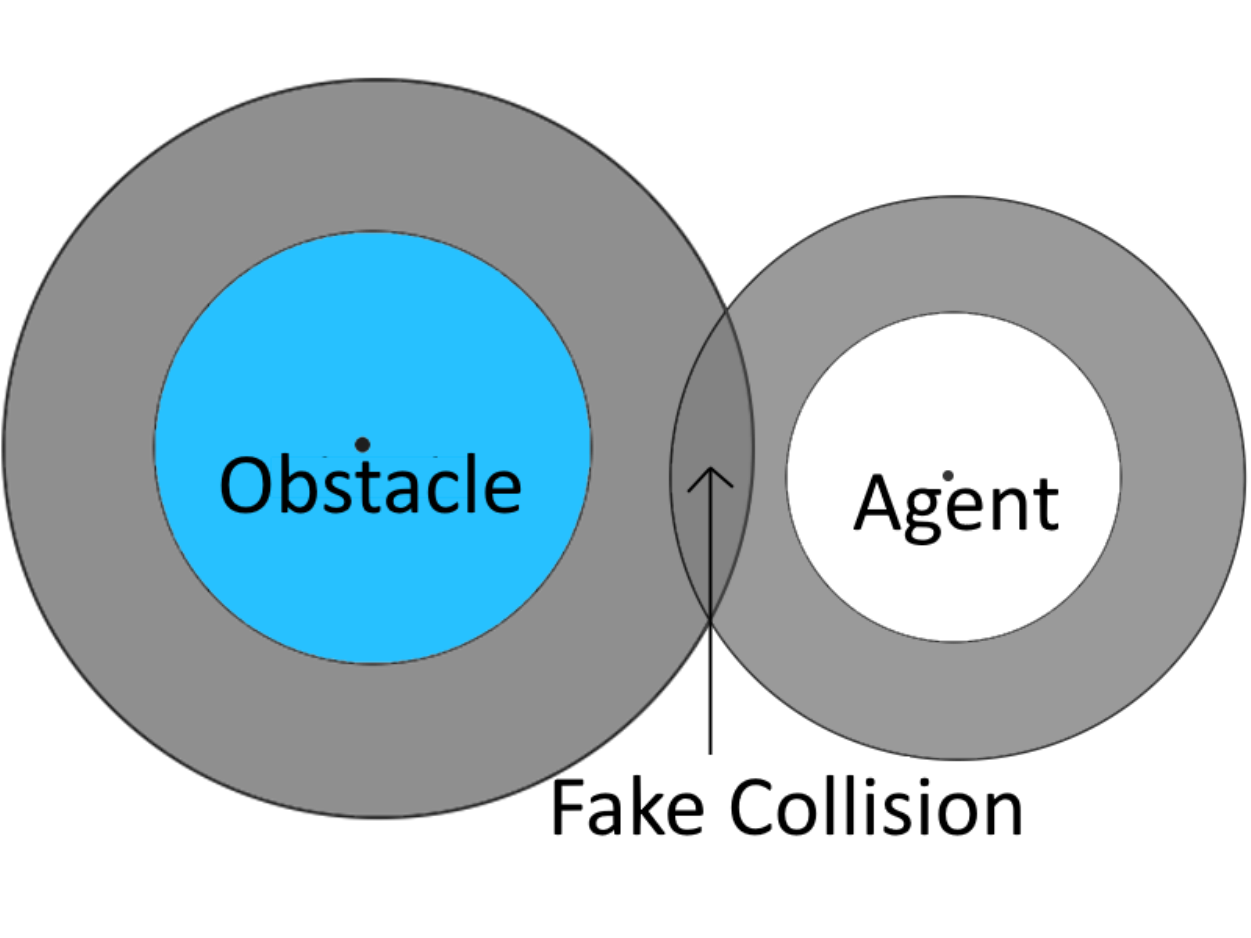}
     \caption{Fake collision}
   \label{fig:2c}
\end{subfigure}
\begin{subfigure}[H]{.241\textwidth}
    \centering
\includegraphics[width=\textwidth]{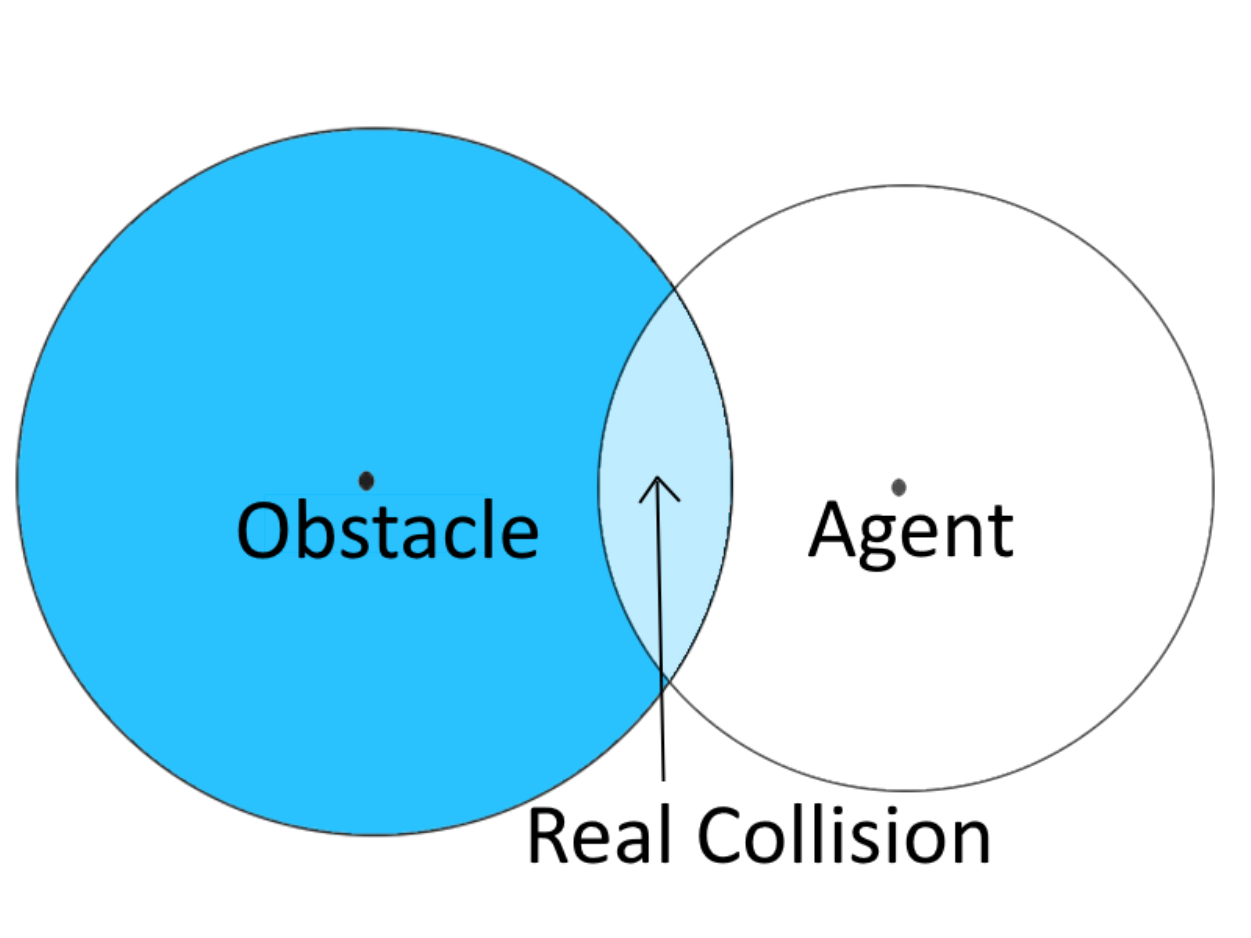}
     \caption{Real collision}
   \label{fig:2d}
\end{subfigure}
\caption{Agent-obstacle collision}
 \label{ag_ob_col}
\end{figure*}

\begin{comment}
\centering{
 \begin{subfigure}{0.3\textwidth}
     \includegraphics[width=\textwidth]{image/delta_o.pdf}
     \caption{$\delta_O$}
     \label{delta_o}
 \end{subfigure}
 %\hfill
 \begin{subfigure}{0.3\textwidth}
     \includegraphics[width=\textwidth]{image/sigma_o.pdf}
     \caption{$\sigma_O$}

 \end{subfigure}
 
 \medskip
 \begin{subfigure}{0.3\textwidth}
     \includegraphics[width=\textwidth]{image/ob_fc.pdf}
     \caption{Fake Collisions}

 \end{subfigure}
 %\hfill
 \begin{subfigure}{0.3\textwidth}
     \includegraphics[width=\textwidth]{image/ob_rc.pdf}
     \caption{Real Collision}

 \end{subfigure}
}
 \caption{Agent-Obstacle Collision}
 \label{ag_ob_col}

\end{comment}

\item{Reward Based on Vehicle-Obstacle Collision:}
Another reward component, $r_3$ is taken that considers the collision between agents and obstacles. It works similarly to the agent-agent collision component where, like agents, obstacles also have a danger zone area $\sigma_O$ around them, as depicted in Fig. \ref{ag_ob_col}. The maximum distance between an agent and an obstacle when a collision happens is $\delta_O$. When danger zones overlap, i.e. the distance between an agent and obstacle, $d_O < (\delta_O + \sigma_O)$, a negative reward is added to the agent colliding. This value also increases as the length of the overlapping area increases. This reward value. $r_3$ can be determined as follows,
\begin{equation}
    r_3 = r_3 - T_3(\delta_O + \sigma_O - d_O)
\end{equation}

    \end{enumerate}
After calculating each of the reward components for each UGV agent, their sum is the final guiding reward for navigating the agents through obstructed terrain to reach targets. The reward function for the UGV, $r_G$ can be estimated as follows,
\begin{equation}
    r_G = r_1 + r_2 + r_3 
\end{equation}

This reward is assigned to each UGV using the MADDPG framework.
    \item \textit{UAV Reward Components}
    
The training environments are divided based on the type of vehicle that is being trained. However, due to the similarities in the training environment of UAVs and UGVs, the reward structure for the UAVs is almost similar to the UGV's reward structure. The UAV agent reward is divided into some components. The description and the method of calculation of these components are given below.
    \begin{enumerate}
        \item{Reward Based on Distance to Targets:} Similar to the UGV environment, we have a target encouragement reward component $r_4$ which sums the distance of each target, $p \in \mathcal{P}$ to its closest agent. This distance value is converted to a negative reward which forms the reward component. It should be noted that for the UAV scenario, when a UAV reaches its assigned target, $r_4$ becomes zero and the rest of the reward components still remain active.

\begin{equation}
r_4 = r_4 + T_1( - min(d_p)), \quad \forall p \in \mathcal{P} 
\end{equation}

        \item{Reward Based on Agent-Agent Collision:} Agent-agent collision is more severe for aerial vehicles as they are more sensitive to collision than ground vehicles. For this reason, the same danger zone concept is adopted from UGVs, where each agent has a danger zone, $\sigma_V$ around it, and when two danger zones overlap or cause a fake collision, a negative reward component $r_5$ is added to the agent reward. This component helps the UAVs learn to avoid colliding with each other while navigating. The reward component, $r_5$ can be determined as follows, 

\begin{equation}
    r_5 = r_5 - T_2(\delta_V + \sigma_V - d_V)
\end{equation}

        \item{Reward Based on Agent-Obstacle Collision:} In the obstructed environment, the UAVs also need to navigate through obstacles to reach the targets safely. For this, a reward component $r_6$ has been used similar to the UGV environment of danger zone $\sigma_O$ and a negative reward based on the overlapping area of agents and obstacles is provided, which can be calculated as follows,
\begin{equation}
    r_6 = r_6 - T_3(\delta_O + \sigma_O - d_O)
\end{equation}

        \item {Reward for Returning to Nearest UGV:} Although both of the environments have very similar functions, a key difference in the UAV environment is that the UAVs need to return to their nearest UGV after reaching their targets. To encourage such behavior, another reward component $r_7$ is added for the UAV. Here, if a UAV has not reached its assigned target, it receives a negative reward of magnitude $r_T$. After reaching the target, this reward gets replaced with a reward value based on the distance $d_h$ between the UAV and its nearest UGV.

\begin{equation}
    r_7 = r_7 - T_1(d_h)
\end{equation}

\end{enumerate}
The final UAV reward is calculated by adding each of the reward components together and the reward function $r_A$ for each agent can be determined as follows,
\begin{equation}
    r_A = r_4 + r_5 + r_6 + r_7
\end{equation}

After calculating, we use this reward to train the UAVs.
\end{enumerate}

\subsubsection{MAPPO Framework}

\begin{algorithm}[!h]
\caption{MAPPO pseudo-code}
\label{algoMAPPO}

\begin{algorithmic}[1]
\STATE Initialize $\theta$, the parameters for policy $\pi$ and $\phi$, parameters for critic $V$
\STATE Set learning rate $\alpha$
\WHILE{$step \leq step_{max}$}
    \STATE Set data buffers $D$ as empty list
    \FOR{$i = 1 \to batch\_size$}
        \STATE Set $\tau$ as empty list
        \STATE Initialize $h_{0, \pi}^{(1)},\dots h_{0,\pi}^{(n)}$ actor RNN states
        \STATE Initialize $h_{0, V}^{(1)},\dots h_{0,V}^{(n)}$ critic RNN states
        \FOR{$t = 1 \to T$}
            \FOR{all agents $a$}
                \STATE $P_t^{(a)}, h_{t, \pi}^{(a)} = \pi(o_t^{(a)}, h_{t-1, \pi}^{(a)}; \theta)$
                \STATE $u_t^{(a)} \sim p_t^{(a)}$
                 \STATE $v_t^{(a)}, h_{t, V}^{(a)} = V(s_t^{(a)}, h_{t-1, V}^{(a)}; \phi)$
            \ENDFOR
            \STATE Execute actions $u_t$, observer $r_t, s_{t+1}, o_{t+1}$
            \STATE $\tau + = [s_t, o_t, h_{t, \pi}, h_{t, V}, u_t, r_t, s_{t+ 1},o_{t+1}]$
        \ENDFOR
        \STATE Compute advantage estimate $\hat{A}$ via GAE on $\tau$, using PopArt
        \STATE Compute reward-to-go $\hat{R}$ on $\tau$ and normalize with PopArt
        \STATE Split trajectory $\tau$ into chunks of length $L$
        \FOR{$l = 0 \to T/L$}
            \STATE $D = D \cup (\tau [l : l+T], \hat{A}[l : l+ L], \hat{R} [l : l + L])$
        \ENDFOR
    \ENDFOR
    \FOR{mini-batch $k = 1 \to K$}
        \STATE $b \xleftarrow{}$ random mini-batch from $D$ with all agent data
        \FOR{each data chunk $c$ in the min-batch $b$}
            \STATE update RNN hidden states for $\pi$ and $V$ from first hidden state in data chunk
        \ENDFOR
    \ENDFOR
    \STATE Adam update $\theta$ on $L(\theta)$ with data $b$
    \STATE Adam update $\phi$ on $L(\phi)$ with data $b$
\ENDWHILE

\end{algorithmic}
\end{algorithm}

As a form of reinforcement learning method, Deep Q-Network or DQN became very popular. It was a simple and easy approach to train models to run in simulations in an effective way. But, it also has a fair amount of drawbacks. Notably, DQN tends to be unstable. Being an offline method and policy iteration method, DQN is very data efficient. But because of the way it does this, it is unstable and it cannot learn some difficult environment. One of the alternative solutions to this was the application of policy gradient methods. These methods work by calculating an estimator of the policy gradient and using the value in a stochastic gradient ascent algorithm. So, if you have a policy $\pi_\theta$, based on the reward signal, some direct loss $L$ and propagate it back directly to the network. To overcome the instability, multiple steps of optimization on $L$ using the same sample or trajectory were suggested. However, this was not justified and it often caused large policy updates which were based on a small amount of data.
Trust Region Policy Optimization or TRPO was an alternative approach that tried to deal with the issue of instability, while not sacrificing data efficiency.

\begin{comment}
Here, maximizing an objective function is constrained by the size of the policy update.
\begin{equation}\label{eqTRPOmax}
    \max_\theta \hat{\mathbb{E}}_t \left[ \frac{\pi_\theta (a_t | s_t)}{\pi_{\theta_{old}} (a_t | s_t)}\hat{A}_t \right]
\end{equation}
subject to
\begin{equation}\label{eqTRPOlimit}
     \hat{\mathbb{E}}_t \left[ KL[\pi_{\theta_{old}}] (.|s_t), \pi_\theta (.|s_t) \right] \leq \delta
\end{equation}

Here, $\pi_{\theta_{old}}$ is the old policy and after updating, we get the new policy, $\pi_\theta$. Equation \ref{eqTRPOmax} is used for maximizing the ratio of the new policy by the old policy and multiplying it with the advantages $\hat{A}_t$ which basically means how much a certain action is than the actual on-policy action. What \ref{eqTRPOmax} is essentially trying to do is change the network in a way such that, the policy is trying to maximize the advantage.
\ref{eqTRPOlimit} ensures that the policy is not changed too much, specifically, not more than $\delta$. Here, $KL$ divergence is used to measure the difference between two distributions and it is used in \ref{eqTRPOlimit} to limit the change of the new policy of the network.
\end{comment}

$\pi_{\theta_{old}}$ is the old policy and after updating, we get the new policy, $\pi_\theta$. The ratio of the new policy is maximized by the old policy and is multiplied by the advantages $\hat{A}_t$ which basically means how much a certain action is than the actual on-policy action. This is essentially trying to change the network in a way such that, the policy tries to maximize the advantage.
TRPO worked really well in a lot of situations, but its main drawback is that it is very complicated. It was difficult to implement and not compatible with architectures that included noise or parameter sharing. This led to the proposition of Proximal Policy Optimization or PPO which had the data efficiency of TRPO, while being easier and quicker to understand and implement. 
\begin{equation}\label{eqPPOPR}
    r_t(\theta) = \frac{\pi_\theta (a_t | s_t)}{\pi_{\theta_{old}} (a_t | s_t)}
\end{equation}
$r_t(\theta)$ in Eq. \eqref{eqPPOPR} is called the probability ratio which is used to simplify the policy ratio. TRPO tries to maximize the objective:
\begin{equation}\label{eqLCPI}
    L^{CPI}(\theta) = \hat{\mathbb{E}}_t \left[ \frac{\pi_\theta (a_t | s_t)}{\pi_{\theta_{old}} (a_t | s_t)}\hat{A}_t \right] = \hat{\mathbb{E}}_t \left[ r_t (\theta) \hat{A}_t \right]
\end{equation}

Maximizing $L^{CPI}$ in Eq. \eqref{eqLCPI} without constraints would create excessively large policy updates. To penalize policy changes that move $r_t({\theta})$ away from 1, Eq. \eqref{eqLCLIP} was proposed as the main objective:

\begin{equation}\label{eqLCLIP}
    L^{CLIP}(\theta) = \hat{\mathbb{E}}_t \left[ \min(r_t (\theta)\hat{A}_t, clip(r_t (\theta), 1- \epsilon, 1 + \epsilon)\hat{A}_t) \right].
\end{equation}
\par The main term in Eq. \eqref{eqLCLIP} has two parts. The first term is the Eq. \eqref{eqLCPI}, while the second term is the $clip(r_t (\theta), 1- \epsilon, 1 + \epsilon)\hat{A}_t$ which modifies the objective by clipping the probability ratio. This removes the incentive for the policy to move beyond the factor of $\epsilon$, not letting it move too far and cause instability. The next part is the minimum function. What it does is take the lower bound of what we know is possible. As we can have multiple values for which the network can learn, if anything but the lower bound of the possible values is selected, there is a higher chance for the learning to collapse or be unstable. Therefore, selecting the absolute minimum value ensures the safest form of update possible. 
\par The form of MAPPO used in this experiment is very similar to the structure of the single-agent PPO but modified for multiple agents. It works by learning a policy $\pi_\theta$ and a value function $V_\phi(s)$. Separate neural networks are used to represent these functions where variance reduction is performed using $V_\phi(s)$. As it is only used during training, it can use global information not observable to the agent to allow PPO in multi-agent domains to follow the centralized training decentralized execution structure. Here, PPO with a centralized value function is regarded as MAPPO, and in our experiment, MAPPO operates in settings where agents share a common reward for cooperative settings. As our individual training environments are comprised of homogeneous agents, parameter sharing is used to improve the efficiency of learning. PPO has also utilized the decentralized partially observable Markov decision processes (DEC-POMDP) with shared rewards. This is defined by $\langle S, A, O, R, P, n, \gamma \rangle$. $o_i = O(s: i)$ is the local observation for agent $i$ at gloabal state $s$. The transition probability from $s$ to $s'$ is denoted by $P(s'|s, A)$ for the joint action $A = (a_1, \dots, a_n)$ for $n$ agents. Action $a_i$ is produced from local observation $o_i$ by agent using policy $\pi_\theta (a_i|o_i)$, parameterized by $\theta$ and jointly optimize the discounted accumulated reward,
\begin{equation}
J(\theta) = \mathbb{E}_ {A^t,s^t} [\sum_t \gamma^tR(s^t, A^t)]
    \end{equation}

\subsubsection{MAPPO Reward Calculation}
As both MAPPO and MADDPG use the same environment, the reward components used in MAPPO are the exact same components used in MADDPG which can be found in the MADDPG Reward Component section \ref{rewardcalc}.

\section{Experimental Evaluation}\label{experiment}
This section describes the details of the simulation platform, the training parameters used for training, and the performance metrics used for evaluation along with simulation results and analysis. We have evaluated the performance of different methods: Simplified IADRL (Baseline) \cite{qie2019joint}, MEANCRFT-MADDPG, MEANCRFT-MAPPO, and CRFT w/o Zones (does not apply modified mean-shift algorithm-based zoning heuristics).

%\label{chapter_experimentalResults}
\subsection{Simulation Platform}
We have created the simulation training environment based on OpenAI's Gym platform. It has been created to simulate an environment with a fixed number of obstacles where a collaborative coalition of UAV-UGV will be deployed to plan a reasonably safe and fast path toward some targets.
%Every quadrant of the environment is considered 1\si{\km} unit in length, where the origin is centered at 0. The size of the UGVs and UAVs is 1\si{\m}, whereas the size of the targets and obstacles is 2\si{\m} and 3\si{\m}, respectively. For the zone radius, we have used 25\si{\m}, since it is a convenient distance for the size of the vehicles and the obstacles we chose.
We have assumed that the area of our customized environment is $2000 \times 2000 m^2$. The position of agents, targets, and obstacles are distributed randomly following uniform random distribution at the start of each episode to ensure the robustness of our work in varying environments. For the zone radius, we have used 25\si{\m}, since it is a convenient distance for the size of the vehicles and the obstacles we chose.

% Observation space and action space. I'm not sure.

All the agents, targets, and obstacles are spawned in random places at the start of each episode. Essentially, in every episode, a new map is given to the training. We have ensured the randomization of the training process in this way. UGVs navigate on the ground surface to reach their destination. They only require motion along the 2 axes to move on the surface they are on. For this, we have selected a 2D environment for training our UGV agents. To create this environment, we have used the environment structure provided by the Multi-agent Particle Environments(MPE)\cite{MPE} package and created our own custom scenario. It uses the geometric coordinate system where the center of the environment is the origin and the environment extends from -1 to +1 on both axes. The agents and landmark components provided by the package are used to build the environment. Agents are used to represent UGVs while landmarks are used to represent targets, obstacles, and rough terrains. Before each episode, the positions of the agents, targets, and obstacles are randomly generated. A similar environment is used for UAV training. Table~\ref{tab:simulation_params} summarizes the major simulation parameter values. 
%%%%%%%%%%%%%%
%%%%%%%%%%%%%%%%%
\begin{table}[!b]
\centering
\caption{Parameters used in simulation environment}
\begin{tabular}{|lc|}
\hline
Parameters & Values\\
\hline\hline
Environment Size & 2000\si{\m} X 2000\si{\m} \\
Number of Obstacles & 1 to 6\\
Number of UGVs & 1 to 4\\
Number of UAVs & 2 to 16\\
Number of Targets & 3 to 20\\
Speed & 1\si{\ms^{-1}}\\
Zone Radius & 10\si{\m} to 25\si{\m}\\
UAV Flight Range & 50\si{\m}\\
\hline
\end{tabular}
\label{tab:simulation_params}
\end{table}
\subsection{Training Parameters}
For training the models in their respective environments using MADDPG, we need to set some hyperparameters for each type of training. Numerous experiments are conducted to set the suitable values for these hyperparameters. We use the same values for training both models. Here, we use $0.01$ as the learning rate for both the actor and critic network and a batch size of 1024 is considered for sampling the memory buffer. The discount factor $\gamma$ is set to $0.99$ and $0.01$ is used as the value for $\tau$. Table~\ref{tab:training_params} lists the training parameter values of the developed MEANCRFT-MADDG and MEANCRFT-MAPPO systems. 

\begin{table*}[!htb]
\centering
\caption{Training parameters used in MEANCRFT-MADDPG and MEANCRFT-MAPPO}
\begin{tabular}{|ccc|}
\hline
Parameters & MADDPG Values & MAPPO Values \\
\hline\hline
$\alpha$ & 0.01 & /\\
$\beta$ & 0.01 & /\\
$\gamma$ & 0.99 & 0.99\\
$\tau$ & 0.01 & 0.01\\
Activation & / & Tanh\\
GAE $\lambda$ & / & 0.95\\
Gain & / & 0.01\\
Batch Size & 1024 & 25\\
Memory Size & $10^6$ & /\\
Number of First Hidden Layer & 64 & 64\\
Number of Second Hidden Layer & 64 & 64\\
\hline
\end{tabular}
\label{tab:training_params}
\end{table*}

\begin{comment}

\begin{table}[!h]
\centering
\begin{tabular}{|lc|}
\hline
Parameters & Values \\
\hline\hline
$\alpha$ & 0.01\\
$\beta$ & 0.01\\
$\gamma$ & 0.99\\
$\tau$ & 0.01\\
Batch Size & 1024\\
Memory Size & $10^6$\\
Number of First Hidden Layer & 64\\
Number of Second Hidden Layer & 64\\
Steps for Updating Target Networks & 100\\

\hline
\end{tabular}
\caption{Training parameters used in MEANCRFT-MADDPG}
\label{tab:training_params}
\end{table}

\begin{table}[!h]
\centering
\begin{tabular}{|lc|}
\hline
Parameters & Values \\
\hline\hline
$\gamma$ & 0.99\\
$lr$ & $7 \times 10^{-4}$\\
Activation & Tanh\\
Buffer Length & 25\\
Epoch & 10\\
fc Layer Dimensions & 64\\
GAE $\lambda$ & 0.95\\
Gain & 0.01\\
Gradient Clip Norm & 10\\
Mini-batch & 1\\
%Network & $RNN$\\
Number of fc & 2\\
Number of fc after & 1\\
Number of GRU Layers & 1\\
Recurrent Data Chunk Length & 10\\
RNN Hidden State Dimensions & 64\\
\hline
\end{tabular}
\caption{Training parameters used in MEANCRFT-MAPPO}
\label{tab:training_params}
\end{table}
\end{comment}

%\subsubsection{The UAV Environment}
%Similar to UGVs, UAVs need to be able to move on 2 axes to reach the targets in their environment. For this, we created a 2D environment using the same MPE\cite{MPE} package. Just like before, the center of the environment is the origin for the geometric coordinate system and the 2 axes can extend from -1 to +1 from the origin. For the UAVs, we used the agent component while the landmark component was used for both targets and obstacles, all of which were available in the package.

\subsection{Performance Metrics}
To evaluate the performance of the studied systems, the following performance metrics have been used:
\begin{itemize}
    \item \textbf{Completion Rate of Tasks}: In our scenario, when all the available targets in the environment have been reached and the UAVs have reached their closest UGV, we call this a completed task. Task completion rate is the summation of the total number of completed tasks in the combined scenario divided by the total number of evaluated episodes, which can be determined as follows,
    \begin{equation}
        \frac{\sum_{e = 1}^{E} T_e}{E},
        \label{metirc:CRT}
    \end{equation}
where, $T_e$ is task completion value of an episode $e$ and $E$ is the maximum number of episodes. The task completion value of an episode is set to 1 when an agent performs its task, Otherwise, 0. 
    \item \textbf{Collisions Per 1K Episode}: This indicates the total number of agent-agent and agent-obstacle collisions per 1K episodes, which can be estimated as follows, 

    \begin{equation}
        \sum_{e = 1}^{1000} (\alpha_e + \beta_e),
        \label{metric:collision}
    \end{equation}
    where, $\alpha_e$ and $\beta_e$ denote the number of agent-agent and agent-obstacle collisions for an episode $e$, respectively,

    \item \textbf{Number of Steps per Episode}: The average of the number of steps required to finish an episode.

    \item \textbf{Accuracy}: The percentage of targets reached out of the total number of targets in the environment per episode. 
    
    \item Completion Time of Task(T): To measure how well the models are performing, we measure the number of steps to complete the task in each evaluation episode. The lesser the value, the shorter the path and time required to complete the task. If the required time step of an episode $e$ is $\phi_e$
    \begin{equation}
        T = \frac{\sum_{e = 1}^{E} \phi_e}{E}
    \end{equation}
\end{itemize}

We evaluated our combined model for 1,000 episodes in different target clusters with different vehicle combinations. For example, 1 UGV with 2 UAVs, 1 UGV with 3 UAVs etc.

\subsection{Simulation Results and Analysis}
%\label{result_analysis
This section illustrates the graphical collections of the results we have collected from our experiment. We have analyzed the results and explained the behaviors of the coalitions both in training and combined model testing. To investigate the performance of the developed models, we randomly vary the numbers of target count (randomly distributed in the environment), coalition combination, and zone radius along the x-axis, and on the y-axis.

\subsubsection{Performance Analysis for Training}

\begin{figure*}[!htb]
 \begin{subfigure}{0.5\textwidth}
     \includegraphics[width=\textwidth]{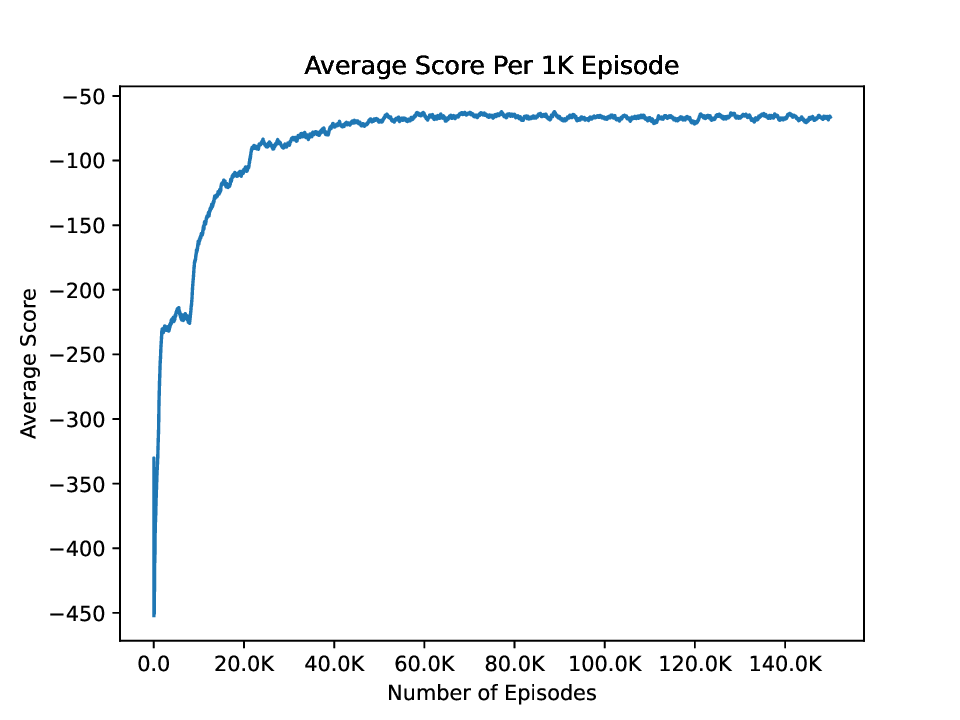}
     \caption{MADDPG average training scores}

 \end{subfigure}
 \hfill
 \begin{subfigure}{0.5\textwidth}
     \includegraphics[width=\textwidth]{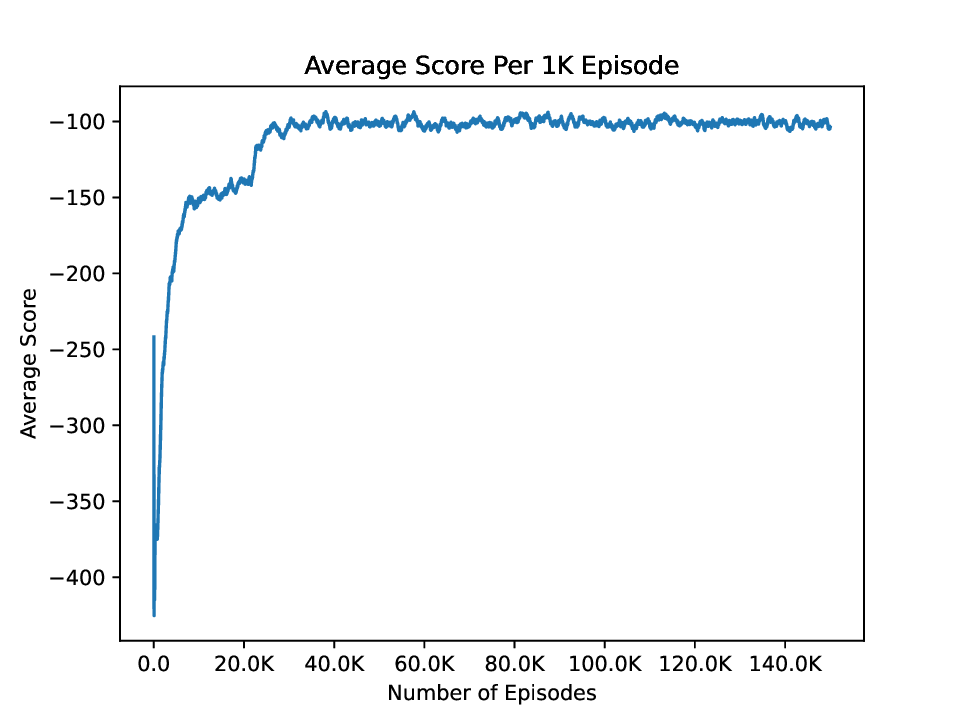}
     \caption{MAPPO average training scores}

 \end{subfigure}

 \caption{Average score per 1000 episodes for UGVs (MADDPG and MAPPO)}
 \label{graph:Score_UGV}

\end{figure*}

Fig. \ref{graph:Score_UGV} shows the reward scores of 2 agents for MADDPG and MAPPO frameworks as the number of training episodes increases. From the graphs, we can conclude that the training process functioned appropriately since the reward converges for both of these cases. Here, we notice that MADDPG's score converged below the score of -50, whereas MAPPO converged below the score of -100. Since both the algorithm environments are the same and use the same reward structure, the MADDPG framework seems to provide a better reward, which implies that the MADDPG framework is able to find shorter or better paths for reaching the target safely.%This indicates that MADDPG performs better in planning a path to reach defined targets.

\subsubsection{Impacts of Varying Number of Targets}
\begin{figure*}[!htb]
 \begin{subfigure}[H]{.48\textwidth}
    \centering
    \includegraphics[width=\textwidth]{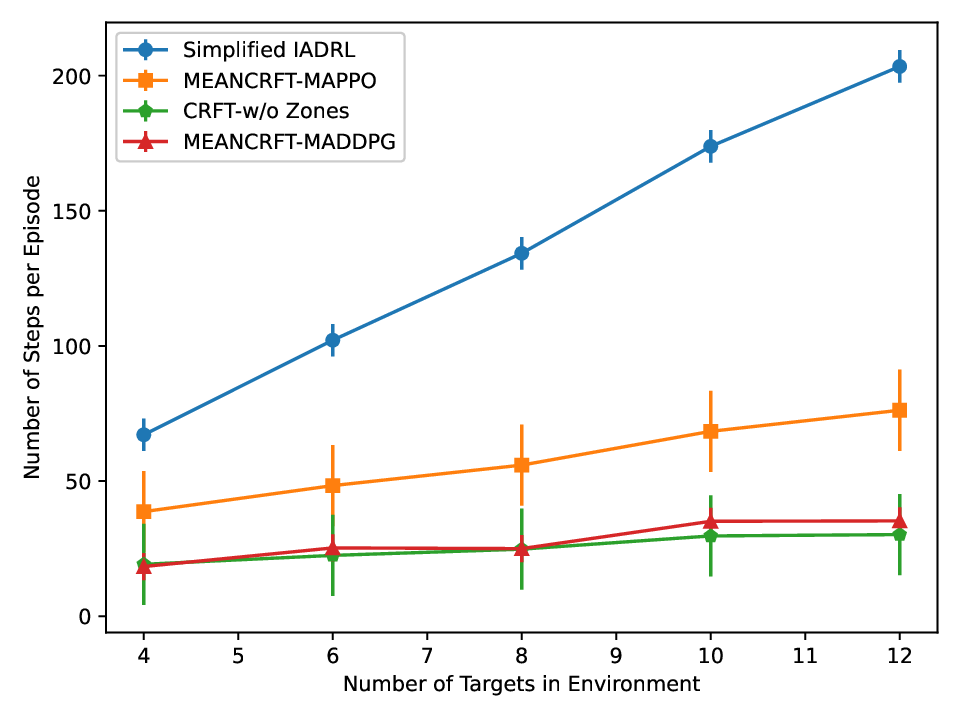}
    \caption{Number of steps per episode}
   \label{steps_gr}
\end{subfigure}
\hfill
\begin{subfigure}[H]{.48\textwidth}
    \centering
\includegraphics[width=\textwidth]{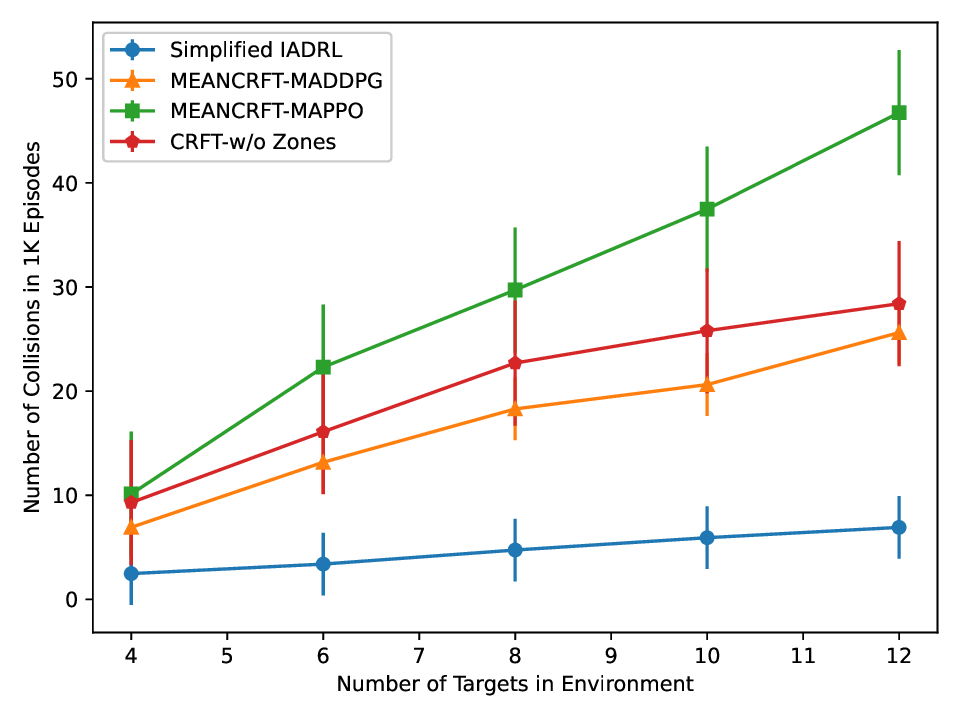}
     \caption{Number of collisions per 1K episodes}
   \label{col_gr}
\end{subfigure}
\hfill
\begin{subfigure}[H]{.48\textwidth}
    \centering
\includegraphics[width=\textwidth]{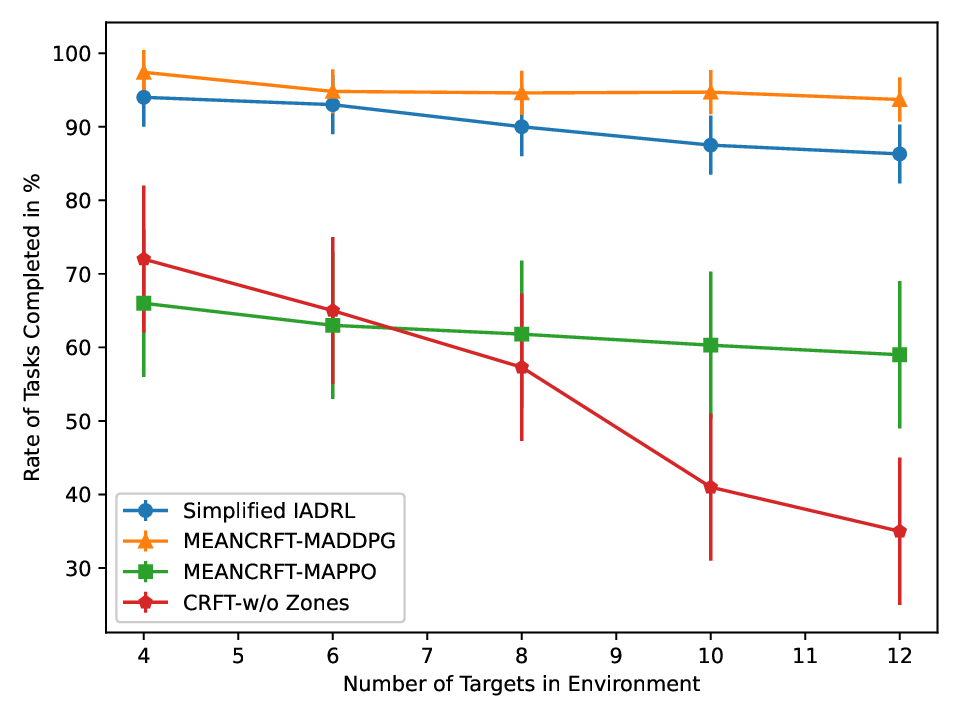}
     \caption{Rate of completed tasks}
   \label{rate_gr}
\end{subfigure}
\hfill
\begin{subfigure}[H]{.48\textwidth}
    \centering
\includegraphics[width=\textwidth]{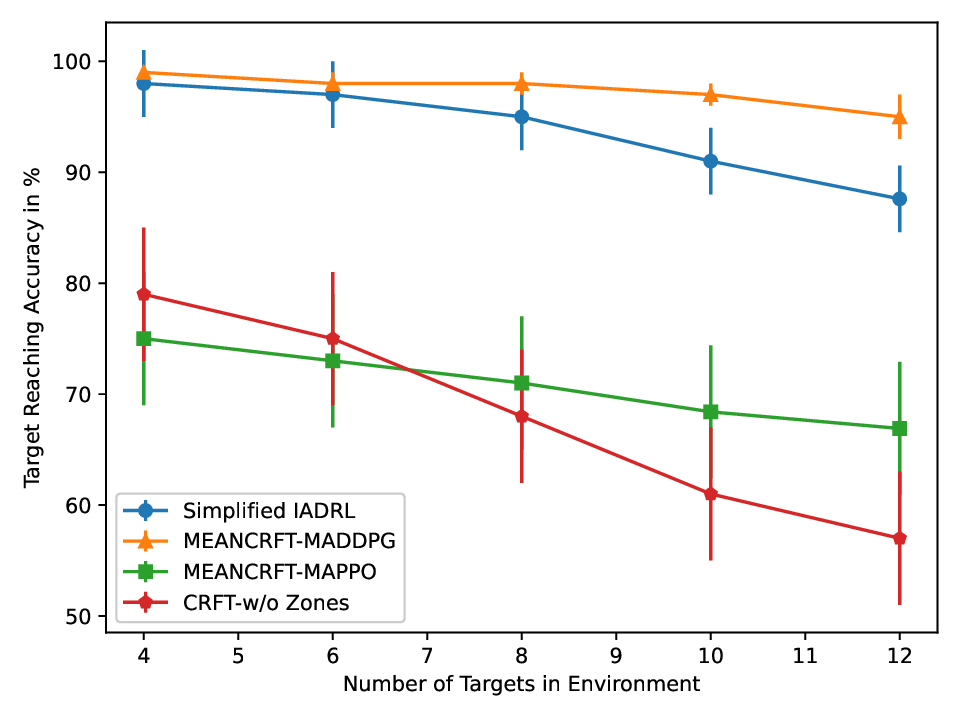}
     \caption{Target reaching accuracy}
   \label{acc_gr}
\end{subfigure}
\caption{Impacts of varying the number of targets}
 \label{target_gr}
\end{figure*}
\begin{figure*}[!ht]
 \begin{subfigure}[H]{.48\textwidth}
    \centering
    \includegraphics[width=\textwidth]{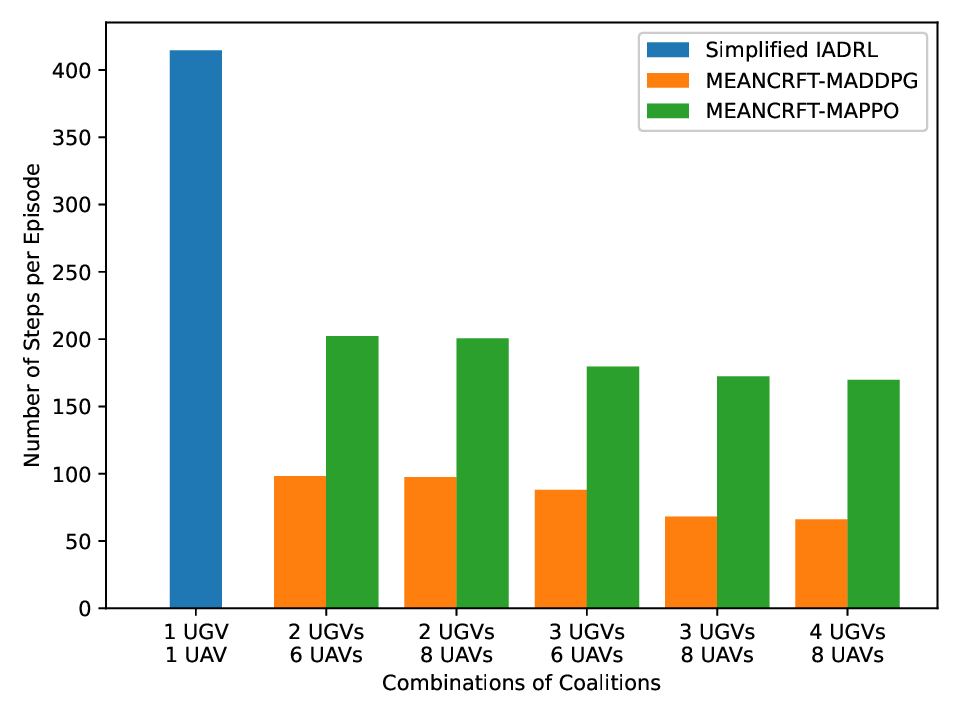}
    \caption{Number of steps per episode}
   \label{steps_bar}
\end{subfigure}
\hfill
\begin{subfigure}[H]{.48\textwidth}
    \centering
\includegraphics[width=\textwidth]{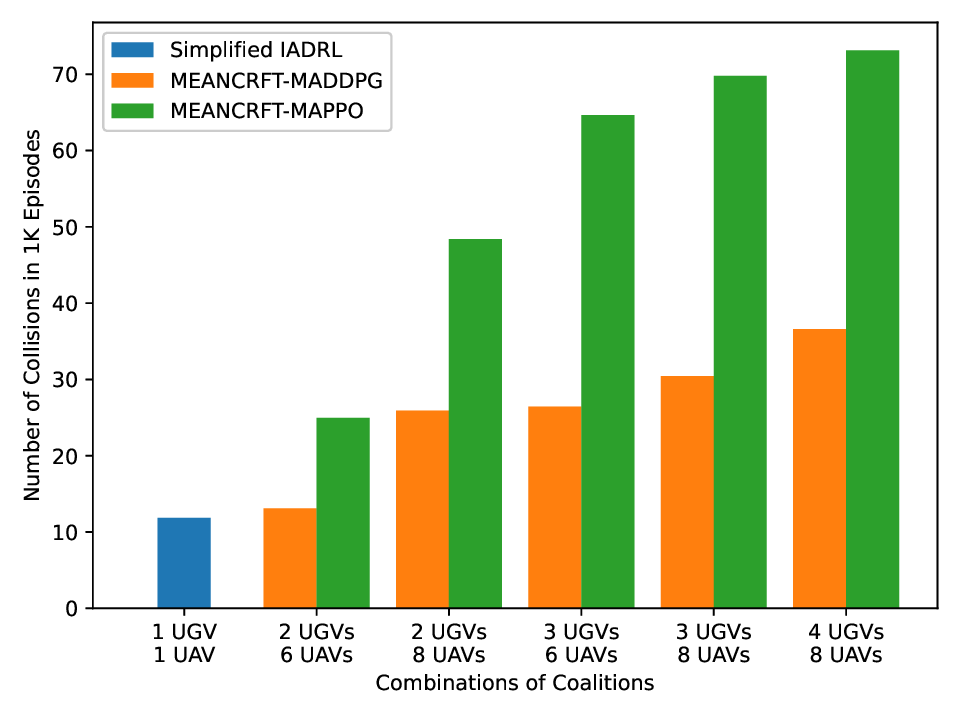}
     \caption{Number of collisions per 1K episodes}
   \label{col_bar}
\end{subfigure}
\hfill
\begin{subfigure}[H]{.48\textwidth}
    \centering
\includegraphics[width=\textwidth]{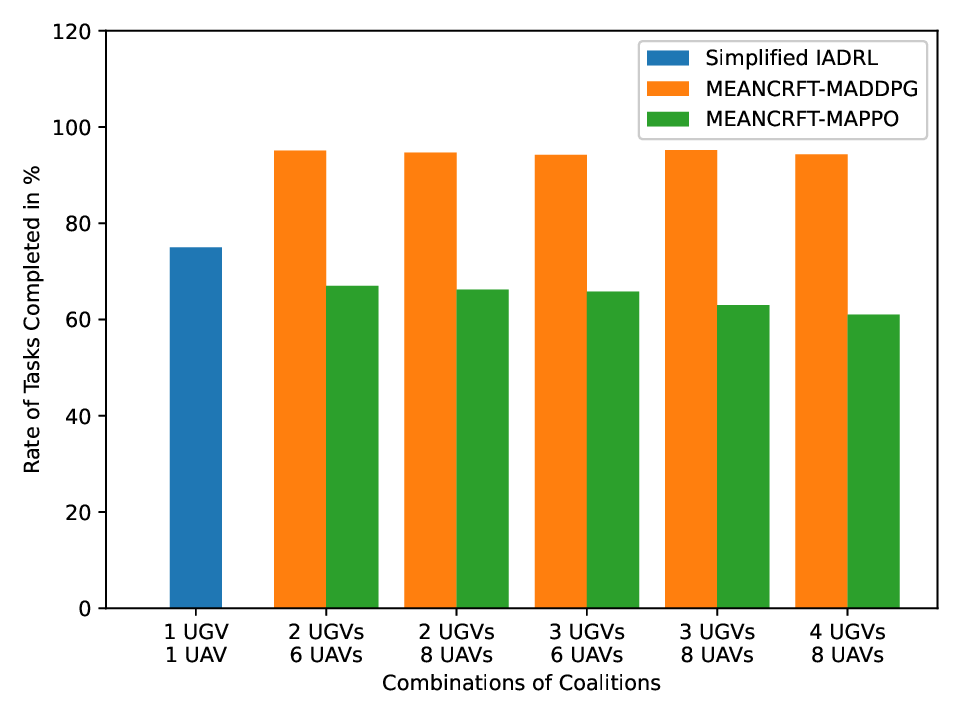}
     \caption{Rate of completed tasks}
   \label{rate_bar}
\end{subfigure}
\hfill
\begin{subfigure}[H]{.48\textwidth}
    \centering
\includegraphics[width=\textwidth]{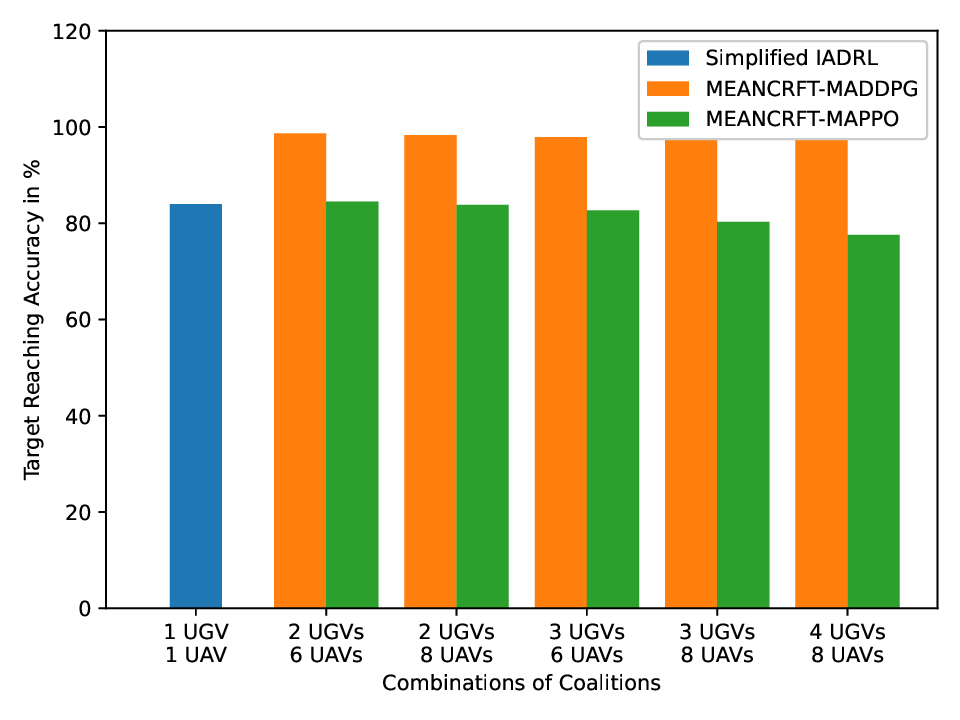}
     \caption{Target reaching accuracy}
   \label{acc_bar}
\end{subfigure}
\caption{Impacts of varying the combination of UAV-UGV coalitions}
 \label{comb_bar}
\end{figure*}
%%%%%%%%%%%
%%%%%%%
Fig. \ref{target_gr} varies the number of targets from 4 to 12, the obstacle count is also varied and measures the performance of the metrics, where, the radius of each zone is chosen as 25m and a total of 2 UGVs and 8 UAVs are deployed in the environment. 

\par Fig.~\ref{target_gr}(\subref{steps_gr}) shows the number of steps required to reach the targets, when the number of targets varies in the environment. Here, we clearly notice that the MEANCRFT-MADDPG approach works considerably better than the baseline IADRL system. Moreover, the graph of CRFT w/o Zoning seems to work just as well as the MEANCRFT-MADDPG one. This is because, to calculate the number of steps per episode, the agents need to complete the episode by reaching all the targets. CRFT w/o Zoning tends to reach all the targets in an episode only when the randomly generated targets are very close to each other and the targets are minimal in number. In that case, the battery limitation issue of UAVs does not harm the performance. In all other cases, the agents fail to reach all the available targets because of the limited step counts which represent the limited battery of a UAV.

\par Fig.~\ref{target_gr}(\subref{col_gr}) demonstrates that the number of collisions per 1000 episodes increases with the increasing number of targets in the environment, and this happens because when the number of targets is minimal, most of the UAVs are not deployed in action since the number of aerial targets is less than the number of available UAVs. Although the number of UAVs and UGVs in the environment is constant, the increase in the number of targets enables the UGVs to deploy a higher number of UAVs to reduce the overall time of the episode. This results in a higher number of collisions with the increase in the number of agents. Simplified IADRL, however, faces few collisions since it has only one UGV and UAV. Here, the MEANCRFT-MADDPG underperforms compared to the IADRL approach because IADRL only uses a single pair of UAV and UGV while MEANCRFT-MADDPG uses multiple UAVs and UGVs which naturally results in a higher number of collisions.

\par Fig.~\ref{target_gr}(\subref{rate_gr}) illustrates the rate of tasks completed in percentage, which takes into account only those episodes where all the targets are reached and all the UAVs have come back to the stationary UGV, and only then it is deemed as a completed task. Here, the without zoning approach fails miserably, since with the increase in the number of targets, the method fails to determine how to get the deployed UAVs back to their respective UGV which might be very far, which results in the UAVs losing battery and failing to complete the task. The MEANCRFT-MADDPG works exceptionally well, with an almost perfect rate of success for the lower number of targets, which decreases a bit, but still retains an impressive value when the target count gets much larger. The MEANCRFT-MAPPO fails substantially as the number of targets increases. In our experiments, we realized that MEANCRFT-MAPPO's on-policy approach works well when the environment is simple and straightforward. As the agent count and complexity of the environment increases, it starts to perform worse. In the developed system, we have tried to fine-tune the environment and other parameters; however, it is not significant enough compared to MEANCRFT-MADDPG.

\par Fig.~\ref{target_gr}(\subref{acc_gr}) depicts the target reaching accuracy, which is very similar to the previous metric. The difference is that we consider reaching the individual targets, not completed tasks like the previous metric. Here, we count the ratio of the number of targets reached by a method to the total number of targets. The IADRL method tries to maintain its high accuracy initially, but its accuracy dwindles as the number of targets in the environment increases. The MEANCRFT-MADDPG always maintains a higher accuracy than the IADRL and this difference is much more noticeable as the target count increases. Its performance does go down slightly as more targets are added, but it still maintains an accuracy above ninety-five percent. This shows that the developed MEANCRFT-MADDPG approach is robust against changes in target counts compared to conventional approaches. Because of the higher complexity of the environment, MEANCRFT-MAPPO underperforms, and its accuracy also decreases as the target count increases. CRFT w/o Zones starts with an eighty percent accuracy, but because of its limited reachability, its performance drastically falls as it fails to reach most of the targets.
%%%%%%%%%%%%%%
%%%%%%%%%%%%%
\begin{figure*}[!ht]
 \begin{subfigure}[H]{.48\textwidth}
    \centering
    \includegraphics[width=\textwidth]{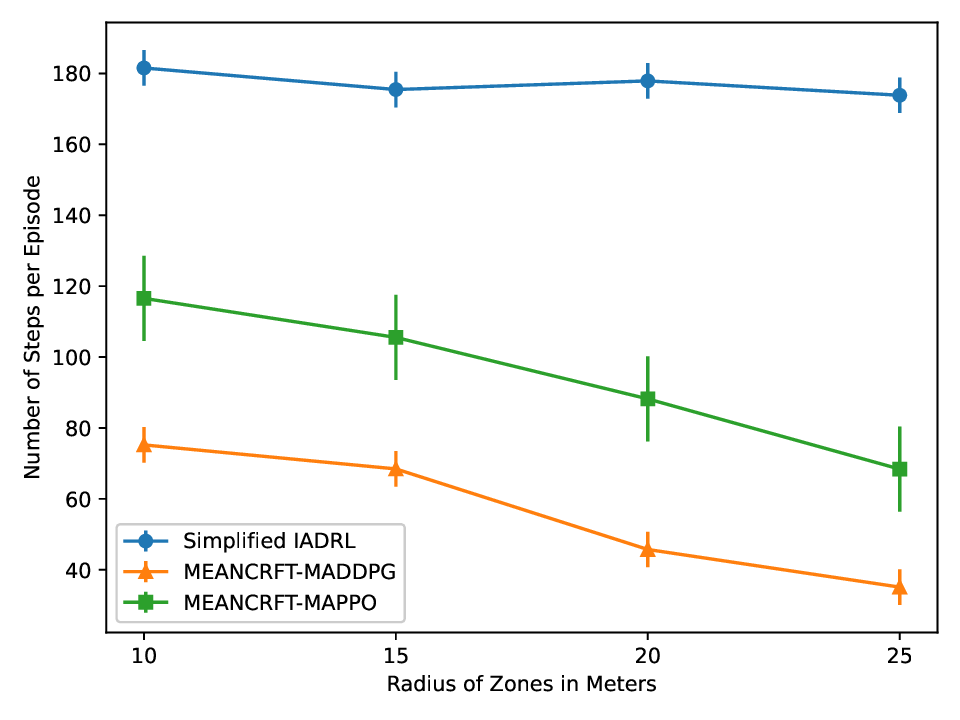}
    \caption{Number of steps per episode}
   \label{steps_rad}
\end{subfigure}
\hfill
\begin{subfigure}[H]{.48\textwidth}
    \centering
\includegraphics[width=\textwidth]{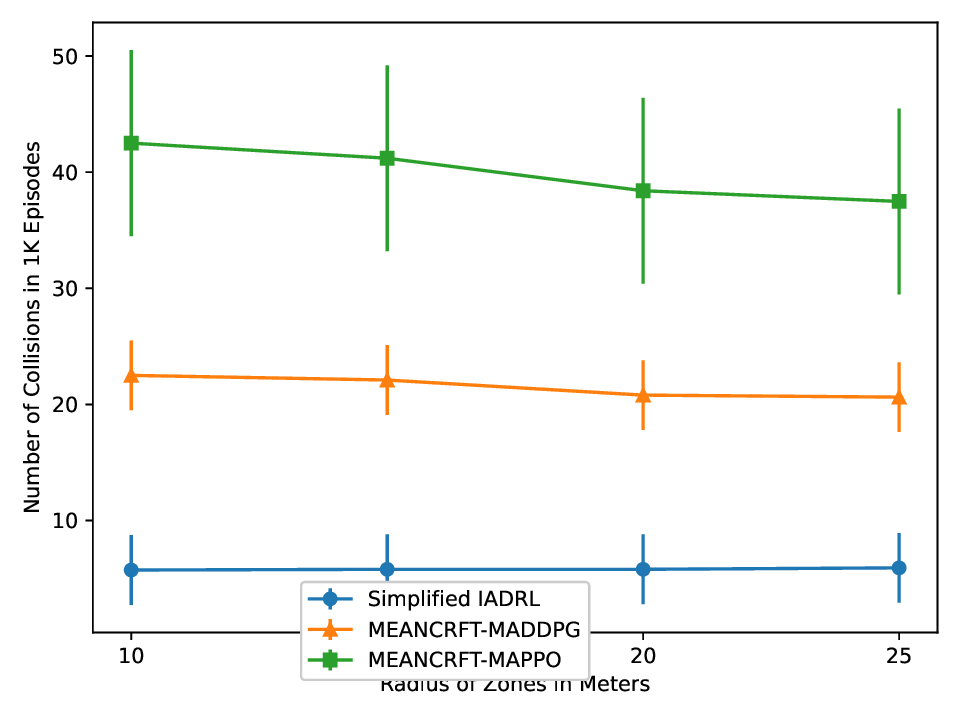}
     \caption{Number of collisions per 1K episodes}
   \label{col_rad}
\end{subfigure}
\hfill
\begin{subfigure}[H]{.48\textwidth}
    \centering
\includegraphics[width=\textwidth]{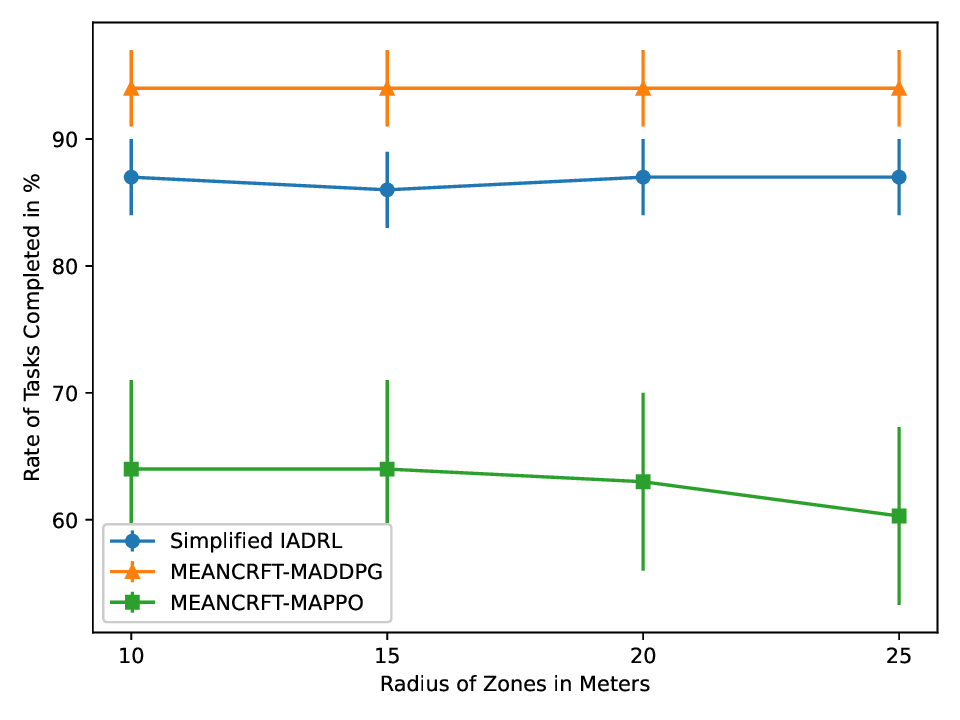}
     \caption{Rate of completed tasks}
   \label{rate_rad}
\end{subfigure}
\hfill
\begin{subfigure}[H]{.48\textwidth}
    \centering
\includegraphics[width=\textwidth]{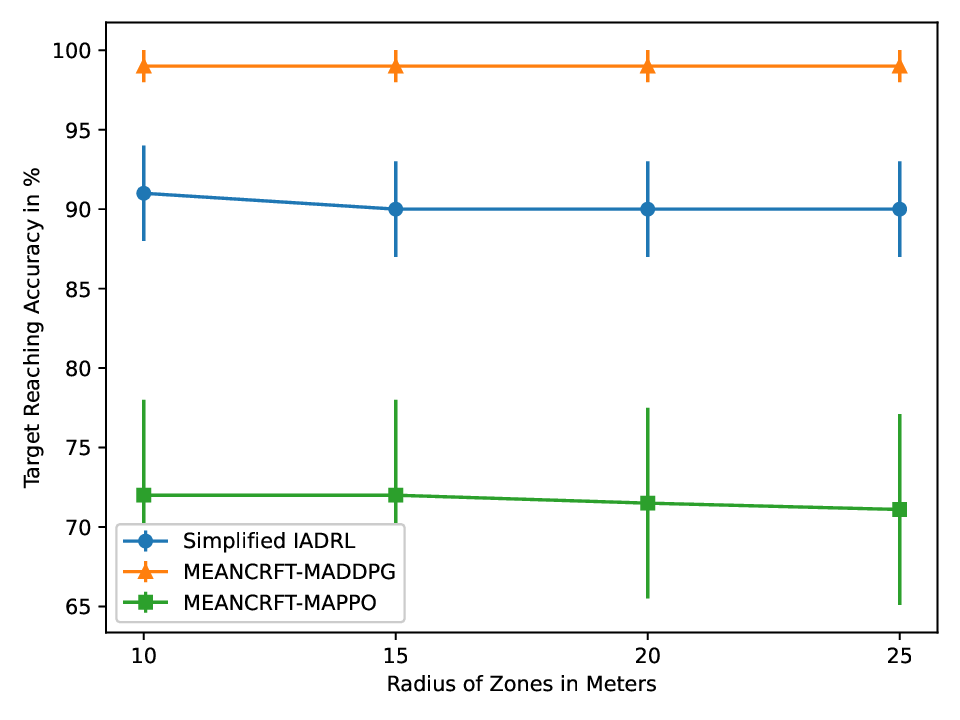}
     \caption{Target reaching accuracy}
   \label{acc_rad}
\end{subfigure}
\caption{Graphs varying the radius of zones}
 \label{comb_rad}
\end{figure*}
\subsubsection{Impacts of Varying the Combination of UAV-UGV Coalitions}

\par Fig.~\ref{comb_bar} varies the combination of UAV-UGV coalitions and measures the performance of the metrics, where the radius of the zone is chosen as 25m, the target count is set to 10, and the obstacle count can vary. The combinations tested in the experiment are 1-1, 2-6, 2-8, 3-6, 3-8, and 4-8, where the first digit indicates the number of UGVs and the second digit is the number of UAVs. As the simplified IADRL method considers only one 1 UAV and 1 UGV, therefore, result analysis in Fig.~\ref{comb_bar} in case of more than 1 UAV and 1 UGV for IADRL has not been shown. 

\par Fig.~\ref{comb_bar}(\subref{steps_bar}) shows the steps required for each method when the number of coalitions varies in the environment. Here, the general trend is that as agent count increases, the number of time steps decreases, which is self-explanatory as more agents naturally get the job done faster. The developed MEANCRFT-MADDPG model seems to work much better than the simplified IADRL or MEANCRFT-MAPPO model in all cases. Since MEANCRFT-MADDPG and MEANCRFT-MAPPO both use more than a single UAV-UGV pair, they easily outperformed IADRL. However, because of the increased complexity of the UAV environment, MEANCRFT-MAPPO is not as effective as MEANCRFT-MADDPG in reducing the number of steps needed to reach the targets. 

\par In Fig.~\ref{comb_bar}(\subref{col_bar}), we demonstrate that the number of collisions per 1000 episodes increases with the increasing number of vehicles. The Simplified IADRL faces few collisions since it has only one UGV and UAV, however, MEANCRFT-MADDPG works almost the same with the increasing number of agent counts. The MEANCRFT-MADDPG and MEANCRFT-MAPPO both have an increasing collision count with the increasing number of agents in the environment, but the increase in MEANCRFT-MADDPG is much more forgiving compared to MEANCRFT-MAPPO because of its robustness against complex environments.
%%%%%%%%%%%%%%%%%%%
%%%%%%%%%%%%%%%%%%
\par Fig.~\ref{comb_bar}(\subref{rate_bar}) illustrates the rate of tasks completed with various combinations of UAV-UGV collision. Within the given experiment, the rate of completion remains almost constant as the number of coalitions increases, which means that for the given target count, all the combinations tested have worked reasonably well. Here, MEANCRFT-MADDPG outperforms the other methods by maintaining a rate above 90\%. Fig.~\ref{comb_bar}(\subref{acc_bar}) shows similar results we found in the previous graph. The target reaching accuracy in MEANCRFT-MADDPG is almost 10\% better than the IADRL and is also consistent as the number of coalitions increases. The MEANCRFT-MAPPO maintains its accuracy with IADRL while trying to maintain consistency as the coalition count increases. This shows that both of our multi-agent approaches stay robust with an increase in agent count, and sometimes may even perform better with a higher agent count.

\subsubsection{Impacts of Varying the Radius of the Zone}

\par Fig.~\ref{comb_rad} varies the radius of zones from 10m to 25m in this experiment and measures the performance of the metrics in each case. Here, the target count is fixed at 10, the obstacle count can vary, and 2 UGVs and 8 UAVs are deployed in the environment. 
%%%%%%%%%%%%
%%%%%%%%%%%
\par Fig.~\ref{comb_rad}(\subref{steps_rad}) shows the required number of steps for each of these methods with the varying radii of the zones. When the radius is low, dynamic coalitions work similarly to IADRL where the entire coalition moves to targets individually, or at least as far as possible until the UGVs face obstacles, and then deploy UAVs, and then the UAVs reach the targets and return. Our coalition MEANCRFT-MADDPG works better than the other methods because there are multiple coalitions involved, and multiple targets are being selected at a time. As we increase the radius size, our system MEANCRFT-MADDPG starts to get significantly better. In Fig.~\ref{comb_rad}(\subref{col_rad}), we demonstrate the number of collisions per 1000 episodes. Here, Simplified IADRL has the least collisions because it has only 1 UGV and 1 UAV. In our method MEANCRFT-MADDPG, we notice a small downward trend as the radius increases. We first assumed that the collision count would increase with radius as more agents would be deployed per zone. But, with the limited number of targets in the environment, the coalitions needed less number of steps to reach the targets with the increase in radius size. As a result, the collision count did not vary much. Fig.~\ref{comb_rad}(\subref{rate_rad}) illustrates the rate of tasks completed in percentage. We note that the rate of completed tasks does not change much with the change in radius for MEANCRFT-MADDPG. This shows that our approach is strong against changes in the radius of the zones. MEANCRFT-MAPPO does not perform well because of the increased complexity of the environment. As the number of targets is already high in the environment, it starts to struggle from the beginning and maintains similar performance. We show the target reaching accuracy in Fig.~\ref{comb_rad}(\subref{acc_rad}) which is almost similar to the previous metrics in Fig.~\ref{comb_rad}(\subref{acc_gr}) and Fig.~\ref{comb_rad}(\subref{acc_gr}) when compared to Completion Rate of Tasks.

From the charts and graphs above, we can come to the conclusion that while different metrics show different trends for different configurations, one of our proposed methods MEANCRFT-MADDPG works much better than the baseline, the non-zoning method, and MEANCRFT-MAPPO.

\section{Conclusions}\label{conclusion}
In this work, we proposed a unique method for one-to-many collaboration between a UGV and a few UAVs. Our methodology used customized training for UGVs and UAVs to develop an efficient coalition needed to navigate through an environment filled with obstacles. The key contribution of our work is extending the flexibility of the number of UAVs and UGVs in a coalition and the quick reaching of various target points while avoiding collisions. A distinctive feature of our method is the division of the targets into various circular zones based on density and range by utilizing a modified mean-shift clustering, which was a heuristic we needed to develop to enhance our approach. The result of our numerical experiments showed that, in terms of target navigation time, collision avoidance rate, and task completion rate, the suggested approach performed better or within an acceptable range when compared to the existing solutions. Our future work plan includes implementing this using the MLAgents package of Unity, since Unity has much better built-in properties that mimic the physical world really well, and models trained there tend to be much more realistic.
\section*{Acknowledgement}
This work was supported by the King Saud University, Riyadh, Saudi Arabia, through the Researchers Supporting Project under Grant RSP2023R18.
%\nocite{*}  
\bibliographystyle{elsarticle-num-names}
\bibliography{references}
\end{document}